# Skellam Mixture Mechanism: a Novel Approach to Federated Learning with Differential Privacy

Ergute Bao
National University of Singapore
bao@u.nus.edu

Yizheng Zhu
National University of Singapore
yzhu@nus.edu.sg

Xiaokui Xiao
National University of Singapore
xkxiao@nus.edu.sg

Yin Yang
Hamad Bin Khalifa University
yyang@hbku.edu.qa

Beng Chin Ooi
National University of Singapore
ooibc@comp.nus.edu.sg

Benjamin Hong Meng Tan
A*STAR, Singapore
benjamin_tan@i2r.a-star.edu.sg

Khin Mi Mi Aung
A*STAR, Singapore
mmaung@i2r.a-star.edu.sg

## ABSTRACT

Deep neural networks have strong capabilities of memorizing the underlying training data, which can be a serious privacy concern. An effective solution to this problem is to train models with *differential privacy* (*DP*), which provides rigorous privacy guarantees by injecting random noise to the gradients. This paper focuses on the scenario where sensitive data are distributed among multiple participants, who jointly train a model through *federated learning*, using both *secure multiparty computation* (*MPC*) to ensure the confidentiality of each gradient update, and differential privacy to avoid data leakage in the resulting model. A major challenge in this setting is that common mechanisms for enforcing DP in deep learning, which inject *real-valued noise*, are fundamentally incompatible with MPC, which exchanges *finite-field integers* among the participants. Consequently, most existing DP mechanisms require rather high noise levels, leading to poor model utility.

Motivated by this, we propose *Skellam mixture mechanism* (SMM), a novel approach to enforcing DP on models built via federated learning. Compared to existing methods, SMM eliminates the assumption that the input gradients must be integer-valued, and, thus, reduces the amount of noise injected to preserve DP. The theoretical analysis of SMM is highly non-trivial, especially considering (i) the complicated math of DP deep learning in general and (ii) the fact that the mixture of two Skellam distributions is rather complex. Extensive experiments on various practical settings demonstrate that SMM consistently and significantly outperforms existing solutions in terms of the utility of the resulting model.

PVLDB Artifact Availability:
The source code, data, and/or other artifacts have been made available at https://github.com/SkellamMixtureMechanism/SMM.

## 1 INTRODUCTION

Deep neural networks, especially large-scale ones such as GPT-3 [11], are known for their excellent memorization capabilities [25, 48, 57]. However, it is rather difficult to control what exactly the neural net memorizes, and unintended data memorization can be a serious concern when the underlying training data contains sensitive information [13]. For instance, consider a bank that trains a GPT-like language model on call center transcripts. Due to data memorization, it is possible to extract sensitive information by letting the model auto-complete a prefix, e.g., "my account number is: __". Clearly, if such a model (or its API) is exposed to the adversary, it becomes a ligation machine as attackers can attempt with various prefixes to extract sensitive data, and subsequently sue the bank for privacy violations. Shokri et al. [47] report that simple and intuitive measures often fail to provide sufficient protection, and the only way found to completely address the issue is to train the model with the rigorous guarantees of *differential privacy* (*DP*) [21].

This paper focuses on the scenario that multiple individual participants jointly train a machine learning model using *federated learning* (*FL*) [38] through distributed stochastic gradient descent (SGD) [1, 16, 18, 37]. Specifically, in every iteration, each individual computes the gradients with respect to the current model weights based on her own data; then, gradients from all participants are aggregated to update the model. Note that the gradients from each individual may reveal sensitive information about her private dataset [39, 43, 46, 47, 56]. A common approach to addressing this problem is by employing a *secure multiparty computation* (*MPC*) protocol [5, 8, 14, 17, 26, 30, 55], which computes the aggregate gradients while preserving the confidentiality of the gradients from each individual participant. One advantage of MPC is that it does not require a trusted third party, which can be difficult to establish in some applications, e.g., in finance and healthcare.

Note that although MPC protects individuals' privacy in the gradient update process by concealing the gradient values of each participant, it does not provide any protection against data extraction attacks caused by unintended data memorization [23, 39, 49, 50]. As mentioned earlier, an effective methodology to defend against such attacks is to perturb the gradients to satisfy differential privacy [47]. Since there is no trusted third-party in our setting, such gradient perturbations need to be done in a decentralized fashion. In particular, each FL participant first adds noise to her own gradients; then, the participants collectively aggregate their noisy gradients, through a cryptographically secure protocol, e.g., SecAgg [10], which ensures that (i) the server, who later updates the model based on the aggregated outcome, learns nothing about the perturbed gradients but the outcome itself, and (ii) no participant learns private information about other participants' data, except for the aggregated outcome. Hence, the privacy cost incurred at each iteration only depends upon the sensitivity of the sum of the gradients, as well as the distribution of the aggregated noise. This framework is referred to as distributed differential privacy [27, 31], elaborated in Section 2.4.

Although gradient perturbation under DP has been studied in previous work, it is far from trivial to adapt centralized DP solutions to the distributed setting. For instance, consider the classic DPSGD algorithm [2], in which a centralized party injects random Gaussian noise to the gradient sum in each iteration of the model training process. There are two major challenges for adapting DPSGD to distributed DP. First, the Gaussian distribution is defined over the domain of all real numbers, whereas existing MPC protocols, to our knowledge, require inputs to be represented as *integers* (more precisely, finite field elements) [9, 10, 44]. Although we could sample a real value from the Gaussian distribution, and then quantize the value to an integer, the resulting quantized samples would no longer follow the Gaussian distribution, which, strictly speaking, invalidates the proof in [2] that the method satisfies DP. Second, even if we ignore the privacy risk of using quantized Gaussian samples in DPSGD [2], the privacy analysis of the algorithm also relies on certain mathematical properties of the Gaussian distribution, which no longer hold with quantized samples. Such properties include: (i) the sum of $n$ values sampled from i.i.d. unit-variance Gaussian distribution follows the Gaussian distribution with variance $n$, and (ii) there exists a tight upper bound for the Rényi divergence [41] between two Gaussian distributions. These issues have been largely neglected by earlier distributed DP solutions, e.g., [27, 51, 52].

Recently, three methods [3, 4, 31] were proposed to address the above problem. A common high-level idea of these methods is to require that each participant of FL inject symmetric integer-valued noise (e.g., binomial noise in [4]) to the gradients during each iteration of the training process. Here, the original values of the gradients are assumed to have bounded norms and are integer-valued, elaborated soon. Next, we point out the main drawback of existing methods, which motivates our proposed solution. Recall the two assumption made above: gradients have bounded norms and are integer-valued. While the bounded norm assumption can be enforced by gradient clipping as is done in DPSGD [2], the integer-valued gradients assumption requires a more complicated pre-processing step accompanied by a careful privacy analysis. Here, we briefly explain how existing works enforce this assumption, and we defer a detailed discussion to Section 5. Ref. [4] stochastically rounds the gradients to integers. For example, if $x = \{0.01, 0.01, \ldots, 0.01\} \in \mathbb{R}^d$, then each dimension of $x$ is rounded to 1 with 0.01 probability, and to 0 with 0.99 probability. While the rounded gradient's expectation equals the original one, the norm of the rounded gradient could be significantly larger than the original one. In our example, $x = \{0.01, 0.01, \ldots, 0.01\} \in \mathbb{R}^d$ could be rounded to $\{1, 1, \ldots, 1\}$, causing an almost $\sqrt{d}$ the increase in $\mathcal{L}_2$ norm. Such an increased sensitivity leads to higher amount of perturbations required to satisfy DP, which, in turn, leads to reduced model utility. Ref. [31] proposes a more complicated conditional rounding process to alleviates (to a limited degree) the problem of increased sensitivity, at the expense of introducing additional bias terms to the resulting gradients, as explained in Section 5.

**Remark on integer-valued noises.** A curious reader may wonder, at this point, why we want the clients to inject integer-valued noises in the first place. The motivation is to achieve differential privacy with mechanisms that are implementable on digital computer systems. On the one hand, computers cannot operate over continuous fields directly. Instead, they operate over floating-point numbers with finite precisions. On the other hand, as shown in the seminal paper of Mironov [40], additive DP mechanisms implemented using floating-point arithmetic lead to discrepancies between the actual noise distributions and their analytical forms, which result in privacy violations (the adversary can recover the whole database through making only a few queries and receiving the 'DP' responses). Integer-valued DP noises is a solution to such privacy issue. Using discrete randomness and simple arithmetic operations such as addition and multiplication only, we can sample integer-valued noises whose actual distribution follows their analytical forms. However, existing integer-valued DP mechanisms (e.g., randomized response, binomial mechanism [4], and discrete Gaussian [12]) are not compatible with the framework of DPSGD and distributed DP at the same time. To simply put, the existing mechanisms do not satisfy RDP and the noise distributions are not closed under summation. The contemporaneous work [3] and our solution resolved this problem, at different communication costs and exhibit different privacy-utility trade-offs. We will provide detailed comparisons in Sections 3.4 and 6.

**Our Contributions.** In this work, we propose the *Skellam mixture mechanism* (SMM), a new solution for enforcing distributed differential privacy for federated learning. SMM works by injecting random noise drawn from the mixture of two shifted symmetric Skellam distributions. *Unlike existing solutions,* SMM *does not require its inputs (i.e., the gradients in FL) to be integer-valued.* This eliminates the need for the process of stochastic rounding the gradients, leading to lower noise level required to satisfy DP, and, thus, higher utility of the resulting model. In particular, with carefully selected mixture coefficients and Skellam distribution parameters, SMM produces private and unbiased integer-valued gradient aggregates for updating the model. We prove that SMM satisfies both Rényi-DP and ($\epsilon$, $\delta$)-DP, defined in Section 2. Meanwhile, SMM is compatible with the DPSGD [2] framework and its moment accountant analysis technique, leading to tight bounds on the privacy loss analysis, similar to our competitors [3, 31].

The privacy analysis of SMM is rather challenging, and is a major contribution of the paper. Note that although the Skellam distribution has been used in previous solution [3], the privacy analysis in [3] does not apply to our setting, since the random noise in SMM is sampled from a mixture of two shifted symmetric Skellam distributions, which is more complex than the single Skellam distribution as in [3]. Further, it is unclear how to derive a tight privacy bound for SMM using the results and mathematical tools built in [3]. One major reason is the privacy bound in Ref. [3] for the Skellam noise requires upper-bounding both the $\mathcal{L}_1$ and $\mathcal{L}_2$ sensitivity of the input, and it is unclear how this bound can be extended to the case of a Skellam mixture distribution. To tackle this problem, we first derive a cleaner privacy bound for the Skellam noise that only involves the $\mathcal{L}_2$ sensitivity, which is the foundation of our privacy analysis for SMM. Our analysis technique for the Skellam distribution is of independent interest, and can be applied to the setting of [3] to improve its privacy bounds by removing the dependency on the input's $\mathcal{L}_1$ sensitivity.

We apply SMM to federated learning with distributed SGD, and present the complete training algorithm. As mentioned above, SMM improves model utility by eliminating the step of rounding the gradients, which often significantly increases the sensitivity of the inputs, especially for large models. Extensive experiments using benchmark datasets demonstrate that SMM achieves consistent and significant utility gains over its competitors, under a variety of settings with different privacy and communication constraints.

## 2 PRELIMINARIES

### 2.1 Skellam Distribution

A random variable $Y$ follows a Poisson distribution of parameter $\lambda$ if its probability distribution is $\Pr[Y = k] = \frac{\exp(-\lambda)\lambda^k}{k!}, k = 0, 1, 2, \ldots$. Both the mean and variance of $Y$ is $\lambda$. A random variable $Z$ follows a Skellam distribution if it is the difference between two independent Poisson variables $Y_1$ and $Y_2$. In this work, we restrict our attention to the case where $Y_1$ and $Y_2$ have the same parameter $\lambda$. In this case, the probability distribution of $Z$ is

$$\Pr[Z = k] = \exp(-2\lambda)I_{|k|}(2\lambda), k = 0, \pm 1, \pm 2, \ldots,$$

where $I_v(u) \triangleq \sum_{h=0}^{\infty} \frac{1}{h!\Gamma(h+v+1)} \left(\frac{u}{2}\right)^{2h+v}$ is the modified Bessel function of the first kind. We write that $Z \sim \text{Sk}(\lambda, \lambda)$. By linearity of expectation, $Z$ has mean 0 and variance $2\lambda$.

Skellam distributions have an important property: they are "additive", in the sense that for any two independent Skellam random variables $Z_1 \sim Sk(\lambda_1, \lambda_1)$, and $Z_2 \sim Sk(\lambda_2, \lambda_2)$, their sum $Z_1 + Z_2$ follows a Skellam distribution $Sk(\lambda_1 + \lambda_2, \lambda_1 + \lambda_2)$. This property is crucial in our analysis of the privacy guarantee of the Skellam mixture noise used in our solution.

### 2.2 Rényi Divergence

DEFINITION 1 (RÉNYI DIVERGENCE [53]). *Assuming that distributions $P$ and $Q$ are defined over the same domain, and $P$ is absolute continuous with respect to $Q$, then the Rényi divergence of $P$ from $Q$ of finite order $\alpha \in (0,1) \cup (1, \infty)$ is defined as:*

$$D_\alpha(P \parallel Q) = \frac{1}{\alpha - 1} \log \mathbb{E}_{X \sim P}\left[\left(\frac{P(X)}{Q(X)}\right)^{\alpha-1}\right],$$

*where we adopt the convention that $\frac{0}{0} = 0$ and $\frac{y}{0} = \infty$ for any $y > 0$, and the logarithm is with base $e$.*

We next present some useful properties of Rényi divergence.

THEOREM 1 (CONVEXITY [53]). *For any order $\alpha \in [0, \infty]$ and $0 < \lambda < 1$, Rényi divergence is convex in its second argument. That is, for any probability distributions $P, Q_0$, and $Q_1$*

$$D_\alpha(P \parallel (1-\lambda) \cdot Q_0 + \lambda \cdot Q_1) \le (1-\lambda) \cdot D_\alpha(P \parallel Q_0) + \lambda \cdot D_\alpha(P \parallel Q_1).$$

THEOREM 2 (JOINT QUASI-CONVEXITY [53]). *For any order $\alpha \in [0, \infty]$ and $0 < \lambda < 1$, Rényi divergence is jointly quasi-convex in its arguments, i.e., for any two pairs of probability distributions $(P_0, Q_0)$, and $(P_1, Q_1)$*

$$D_\alpha((1-\lambda) \cdot P_0 + \lambda \cdot P_1 \parallel (1-\lambda) \cdot Q_0 + \lambda \cdot Q_1)$$
$$\le \max\{D_\alpha(P_0 \parallel Q_0), D_\alpha(P_1 \parallel Q_1)\}.$$

### 2.3 Differential Privacy

We say that two datasets $X$ and $X'$ are neighboring if one can be obtained by adding or removing one tuple from the other. The main idea of differential privacy (DP) is to ensure that the outcomes of a randomized mechanism on neighboring datasets are always similar; intuitively, this provides plausible deniability on whether a given data record $x$ belongs to the dataset $X$ or not, and, thus, protects the privacy of the individual whose record is $x$. A classic definition of differential privacy is $(\epsilon, \delta)$-DP [21], as follows.

DEFINITION 2 ($(\epsilon, \delta)$-DIFFERENTIAL PRIVACY [21]). *A randomized mechanism $\mathcal{M}$ satisfies $(\epsilon, \delta)$-differential privacy (DP) if*

$$\Pr[\mathcal{M}(X) \in O] \le \exp(\epsilon) \cdot \Pr[\mathcal{M}(X') \in O] + \delta,$$

*for any set of output $O \subseteq Range(\mathcal{M})$ and any neighboring datasets $X$ and $X'$.*

Note that $(\epsilon, \delta)$-DP can be considered as a worst-case privacy guarantee for a mechanism, as it enforces an upper bound on the probability ratio of all possible outcomes. An alternative definition is Rényi differential privacy (RDP) [41], which is built upon the concept of Rényi divergence, considers the average case privacy guarantee instead.

DEFINITION 3 (RÉNYI DIFFERENTIAL PRIVACY [41]). *A randomized mechanism $\mathcal{M}$ satisfies $(\alpha, \tau)$-Rényi differential privacy (RDP) if $D_\alpha(\mathcal{M}(X) \parallel \mathcal{M}(X')) \le \tau$ for all neighboring datasets $X$ and $X'$.*

Given a function of interest, the canonical way to make it differentially private is to perturb its outcome through noise injection. Specifically, the scale of the noise should be calibrated to the sensitivity of the function of interest [21], formally defined as follows.

DEFINITION 4 (SENSITIVITY). *The sensitivity $S(F)$ of a function $F : \mathcal{D} \to \mathbb{R}^d$, denoted as $S(F)$, is defined as*

$$S(F) = \max_{X \sim X'} \|F(X) - F(X')\|,$$

*where $X \sim X'$ denotes that $X$ and $X'$ are neighboring datasets, and $\|\cdot\|$ is a norm.*

In particular, injecting continuous Gaussian noise sampled from $\mathcal{N}(0, \sigma^2)$ to each dimension of function $F$ satisfies $(\alpha, \frac{\alpha S^2(F)}{2\sigma^2})$-RDP [41], where $S(F)$ stands for the $\mathcal{L}_2$ sensitivity of function $F$. In many applications (e.g., training neural networks with SGD), we also need to analyze the overall privacy guarantee of a mechanism consisting of multiple components. We have the following composition and sub-sampling lemmata for RDP mechanisms.

LEMMA 1 (COMPOSITION OF RDP MECHANISMS [41]). *If mechanisms $M_1, \ldots, M_T$ satisfy $(\alpha, \tau_1), \ldots, (\alpha, \tau_T)$-RDP, respectively, then, $M_1 \circ \ldots \circ M_T$ satisfies $(\alpha, \sum_{t=1}^{T} \tau_t)$-RDP.*

LEMMA 2 (SUBSAMPLING FOR RDP [42, 58]). *Let $M$ be a mechanism that satisfies $(l, \tau(l))$-RDP for $l = 2, \ldots, \alpha$ ($\alpha \in \mathbb{Z}, \alpha \leq 2$), and $S_q$ be a procedure that uniformly samples each record of the input data with probability $q$. Then $M \circ S_q$ satisfies $(\alpha, \tau)$-RDP with*

$$\tau = \frac{1}{\alpha - 1} \cdot$$
$$\log \left( (1-q)^{\alpha-1} (\alpha q - q + 1) + \sum_{l=2}^{\alpha} \binom{\alpha}{l} (1-q)^{\alpha-l} q^l e^{(l-1)\tau(l)} \right).$$

Finally, any mechanism that satisfies $(\alpha, \tau)$-RDP also satisfies $(\epsilon, \delta)$-DP, for values of $\epsilon$ and $\delta$ as follows.

LEMMA 3 (CONVERTING $(\alpha, \tau)$-RDP TO $(\epsilon, \delta)$-DP [12]). *For any $\alpha \in (1, \infty)$, if $D_\alpha(M(X) \| M(X')) \leq \tau$ for any neighboring databases $X$ and $X'$, then $M$ satisfies $(\epsilon, \delta)$-DP for*

$$\epsilon = \tau + \frac{\log(1/\delta) + (\alpha - 1) \log(1 - 1/\alpha) - \log(\alpha)}{\alpha - 1}.$$

## 2.4 Distributed Differential Privacy

The original, centralized differential privacy framework [21] assumes a trusted data curator, who stores the entire private dataset and injects random noise in its response to a query, e.g., the sum query, which computes $\sum_{i=1}^{n} x_i$ given input dataset $X = (x_1, \ldots, x_n)$. The released outcome satisfies (centralized) DP, when the scale of the noise injected is calibrated (by the centralized data curator) to the sensitivity of the query. In this work, we focus on the distributed differential privacy framework [15, 24, 27, 31], which involves multiple participants. Each participant injects a random noise to her own data or query response. After that, all participants collectively run an MPC protocol to amplify the privacy guarantee by hiding the identities of the participants. We follow the same threat model as in previous work [31]. In particular, all participants are honest (i.e., they strictly follow the protocol) but curious (i.e., each participant tries to learn private information from another participant), and it is assumed that no two participants collude. In this paper, we focus on SecAgg [10] as the MPC protocol, which aggregates inputs from participants in a crpytographically secure manner under our threat model. Specifically, SecAgg ensures that no one (including the participants) can infer any information about the private inputs other than its released output. The output of SecAgg should satisfy DP, such that it can be distributed the same way as the result from a centralized DP solution. Accordingly, the scale of the overall noise injected to the data or query result needs to be calibrated to the sensitivity of each participant's input. In other words, distributed DP obtains the same privacy-utility trade-off as in centralized DP setting, without relying on a trusted third party.

## 3 SKELLAM MIXTURE MECHANISM

Section 3.1 formalizes the problem of distributed sum estimation under distributed DP, and Section 3.2 presents the proposed *Skellam Mixture Mechanism* (SMM) for this problem. Section 3.3 presents the foundation of the privacy guarantee of SMM. Section 3.4 establish the privacy and utility guarantees of SMM. Then, in Section 4, we apply SMM to our main problem setting: differentially private federated learning.

## 3.1 Distributed Sum Estimation with Privacy

Suppose that a multi-dimensional dataset $X = (x_1, \ldots, x_n)$ is distributed to $n$ individuals (referred to as *participants* in the following), where participant $i$ possesses data point $x_i \in \mathbb{R}^d$, for $i = 1, \ldots, n$. An un-trusted *server* aims to compute the (approximate) sum of the dataset, i.e., $\bar{x} = \sum_{i=1}^{n} x_i$, from the participants. Agarwal et al. [4] propose a general framework for solving the distributed sum estimation problem with distributed DP. In this framework, each participant first perturbs her data $x_i$ with noise $Z_i$: $x_i^* \leftarrow x_i + Z_i$. Next, a secure aggregation protocol SecAgg [10], run as a black box by the participants, sums up the noisy values $x_i^*$ from all participants, and outputs to the server the result $\bar{x}^* \leftarrow SecAgg(x_1^*, \ldots, x_n^*)$.

According to Ref. [10], SecAgg ensures that no participant (or the server) learns any information about another participant's private data. Hence, it suffices to derive the privacy and utility guarantees of the following mechanism $M$, which injects $n$ independent random noises $Z_i$ to the exact sum:

$$M(x_1, \ldots, x_n) := \sum_{i=1}^{n} x_i + \sum_{i=1}^{n} Z_i.$$

In terms of privacy, we focus on the RDP definition (Definition 3), which can be converted to the classic $(\epsilon, \delta)$-DP (Definition 2) through Lemma 3. In particular, we want that for all neighboring datasets $X, X'$,

$$D_\alpha(M(X) \| M(X')) \leq \tau,$$

for some $\alpha > 1$. We measure the error of $M$ by

$$Err_M = \max_{X \subset \mathbb{R}^d} \frac{1}{d} \mathbb{E} \left\| M(X) - \sum_{x \in X} x \right\|_2^2,$$

where the expectation is taken over the randomness in $M$.

## 3.2 Skellam Mixture Noise

We first consider the case when each participant's data point $x_i$ is one-dimensional. Algorithm 1 shows the pseudo-code of our *one-dimensional Skellam mixture mechanism* (1SMM) for this case. Each participant $i$ first independently flips a coin with heads probability $p_i := x_i - \lfloor x_i \rfloor$ (Lines 2 and 3). If it is tails, then the participant perturbs $\lfloor x_i \rfloor$ with a noise following the Skellam distribution $Sk(\lambda, \lambda)$ (Lines 4 and 5); otherwise, the participant perturbs $\lfloor x_i \rfloor + 1$ (i.e., $\lceil x_i \rceil$) with a noise following the Skellam distribution $Sk(\lambda, \lambda)$ (Lines 6 and 7). Note that, by the definitions of $x_i^*$ and the Skellam distribution, $x_i^*$ is guaranteed to be an integer. Finally, SecAgg aggregates the noisy values from all the participants (Line 9), and the estimated sum $\bar{x}^*$ is released to the server. For the case in which each participant's data point $x_i$ is multidimensional, we simply invoke Algorithm 1 for each dimension independently to obtain a noisy sum of that dimension, as outlined in Algorithm 2.

The result of Algorithm 1 may appear rather difficult to analyze at first sight, as there are $2^n$ possible outcomes of the Bernoulli trials by all participants. An import insight in our analysis is that to derive the utility guarantee of Algorithm 1, it suffices to consider each participant independently. First, note that the perturbed value $x_i^*$ follows a mixture of two shifted symmetric Skellam distributions

**Algorithm 1:** One-dimensional Skellam mixture mechanism (1SMM)

**Input:** A set of private values $\{x_1, \ldots, x_n \mid x_i \in \mathbb{R}\}$.
**Parameters:** Noise parameter $\lambda$.
1 **for** $i \in 1..n$ **do**
2 $\quad p_i = x_i - \lfloor x_i \rfloor$.
3 $\quad$ Sample $y_i$ from a Bernoulli trial with success probability $p_i$
4 $\quad$ **if** $y_i = 0$ **then**
5 $\quad\quad x_i^* \leftarrow \lfloor x_i \rfloor + Sk(\lambda, \lambda)$.
6 $\quad$ **else**
7 $\quad\quad x_i^* \leftarrow \lfloor x_i \rfloor + 1 + Sk(\lambda, \lambda)$.
8 $\bar{x}^* \leftarrow SecAgg((x_1^*, \ldots, x_n^*))$.
**Output:** $\bar{x}^*$.

---

**Algorithm 2:** Multi-dimensional Skellam mixture mechanism (dSMM)

**Input:** A set of private values $\{x_1, \ldots, x_n \mid x_i \in \mathbb{R}^d\}$.
**Parameters:** Noise parameter $\lambda$, data dimension $d$.
1 **for** $i \in 1..n$ **do**
2 $\quad$ **for** $j \in 1..d$ **do**
3 $\quad\quad p_{i,j} = x_{i,j} - \lfloor x_{i,j} \rfloor$.
4 $\quad\quad$ Sample $y_{i,j}$ from a Bernoulli trial with success probability $p_{i,j}$.
5 $\quad\quad$ **if** $y_{i,j} = 0$ **then**
6 $\quad\quad\quad x_{i,j}^* \leftarrow \lfloor x_{i,j} \rfloor + Sk(\lambda, \lambda)$.
7 $\quad\quad$ **else**
8 $\quad\quad\quad x_{i,j}^* \leftarrow \lfloor x_{i,j} \rfloor + 1 + Sk(\lambda, \lambda)$.
9 $\bar{x}^* \leftarrow SecAgg((x_1^*, \ldots, x_n^*))$.
**Output:** $\bar{x}^*$.

---

whose shifted mean values equal $\lfloor x_i \rfloor$ and $\lceil x_i \rceil$, respectively, and the variance of each distribution equals $2\lambda$. In addition, observe that the weights associated with the mixture distributions are $1 - x_i + \lfloor x_i \rfloor$ and $x_i - \lfloor x_i \rfloor$, respectively. Consequently, the expectation of $x_i^*$ equals $x_i$. A corner case is that $x_i$ is an integer. In this case, the perturbed $x_i^*$ can be seen as injecting symmetric Skellam noise $Sk(\lambda, \lambda)$ to $x_i$ itself only. By the linearity of expectation, the expectation of $\bar{x}^*$ also equals $\sum_{i=1} x_i$, i.e., 1SMM yields an unbiased estimator for the sum of private inputs. We present the detailed privacy and utility analysis for 1SMM and dSMM later in Section 3.4.

### 3.3 Skellam Noise Preserves Privacy

Before we analyze the privacy guarantee of SMM, we first show that its building block, i.e., a single symmetric Skellam noise, preserves privacy, formalized as follows.

**Theorem 3** (Rényi divergence of Skellam distributions).
*For any integer $s \in \mathbb{Z}$ satisfying $|s| \leq \Delta_\infty$, any $\alpha > 1$, and any $\Delta_\infty$ satisfying $\alpha < 2\lambda/\Delta_\infty + 1$, we have*

$$D_\alpha(s + Sk(\lambda, \lambda) \parallel Sk(\lambda, \lambda)) \leq \frac{1.09\alpha + 0.91}{2} \cdot \frac{s^2}{2\lambda}. \quad (1)$$

We have the following multi-dimensional extension.

**Theorem 4** (Rényi divergence of multi-dimensional Skellam distributions). *Let $Sk^d(\lambda, \lambda)$ denote a $d$-dimensional variate, where each dimension is independently sampled from $Sk(\lambda, \lambda)$. Then, for any integer-valued vector $s \in \mathbb{Z}^d$ satisfying $\|s\|_2^2 \leq c$ and $\|s\|_\infty \leq \Delta_\infty$, any $\alpha > 1$, and any $\Delta_\infty$ satisfying $\alpha < 2\lambda/\Delta_\infty + 1$, we have*

$$D_\alpha(s + Sk^d(\lambda, \lambda) \parallel Sk^d(\lambda, \lambda)) \leq \frac{1.09\alpha + 0.91}{2} \cdot \frac{c}{2\lambda}. \quad (2)$$

The proof of the above theorem can be found in Appendix C.1 of the technical report version [7]. Next, we highlight the contributions of our theoretical results. First, according to Theorem 3, the privacy guarantee provided by a symmetric Skellam noise of variance $2\lambda$ is comparable (i.e., within a constant factor) with that of adding continuous Gaussian noise of the same variance, which is $\frac{\alpha \cdot s^2}{2 \cdot 2\lambda}$ [41]. Further, as Eq. (1) only involves the quadratic term, the one-dimensional privacy analysis can be easily extended to the multi-dimensional setting by replacing the quadratic term with the squared $\mathcal{L}_2$ norm, as in Theorem 4.

Meanwhile, since additive Skellam noise preserves RDP, by Lemmata 1 and 2, it allows the tight privacy accounting of Skellam noises in applications involving composition and subsampling (e.g., FL), which is elaborated further in Section 4. Note that although our analysis restricts the value of $\Delta_\infty$ to $\Delta_\infty < 2\lambda/(\alpha - 1)$, this constraint only affects the utility of the Skellam noise, not its privacy guarantees. This is because the constraint can be easily enforced by standard $\mathcal{L}_\infty$ clipping. In addition, in the federated learning setting, $\lambda$ is usually much larger than the optimal $\alpha$ (order of RDP) due to the fact that a large number of participants contribute to the overall DP noise, and the optimal $\alpha$ is often relatively small (e.g. less than 10 in our experiments). Hence, the above constraint leads to a sufficiently large range for $\mathcal{L}_\infty$ clipping without causing much utility degradation.

A notable difference between our theoretical result presented in Theorem 4 and the analysis of Skellam noise in [3] is that our result is "cleaner" in the sense that Eq. (2) only involves the $\mathcal{L}_2$ norm (similar to the case of continuous Gaussian noise [41]), whereas the analysis in [3] also involves the $\mathcal{L}_1$ norm of vector $s$. In general, the presence of $\mathcal{L}_1$ sensitivity may lead to an excessive amount of noise for high dimensional data, as the $\mathcal{L}_1$ sensitivity can be $\sqrt{d}$ times larger than the $\mathcal{L}_2$ sensitivity, limiting the applicability of Skellam noises in such applications. Further, the clean bound without the $\mathcal{L}_1$ norm term may also significantly simplify the design of protocols and mechanisms built on top of additive Skellam noises, e.g., algorithms 1SMM and dSMM presented earlier. To avoid the $\mathcal{L}_1$ norm term, we do not use known properties of Rényi divergence, and instead attack the problem directly using basic mathematical tools, which is a novel proving technique of independent interest (presented in detail in Appendix C.1 in the technical report version [7]). In particular, this proving technique leads to long and heavy formulae at the beginning, and yet within a few steps, most terms are canceled out, resulting in a clean bound.

Finally, we mention that there exists an exact sampler for the Skellam distribution, described Appendix A of the technical report version [7]. As a result, adding Skellam noise strictly preserves differential privacy. On the contrary, we are not aware of an exact sampler for the continuous Gaussian distribution. Consequently, the random noise sampled with an inexact sampler only approximately follows the Gaussian distribution; strictly speaking, injecting such noise may violate differential privacy, which is yet another motivation for employing our proposed method that injects Skellam noise.

## 3.4 Theoretical Analysis of Skellam Mixture Mechanism

We present the theoretical analysis of the proposed Skellam mixture mechanism (SMM). The proofs are deferred to Appendix C in the technical report version [7]. In terms of privacy, we have the following theorem for Algorithm 1.

THEOREM 5. *Suppose that each participant's data point $x_i$ satisfies*
$$|x_i|^2 + (|x_i| - \lfloor |x_i| \rfloor) - (|x_i| - \lfloor |x_i| \rfloor)^2 \leq c$$
*and $\lceil |x_i| \rceil \leq \Delta_\infty$. Then, whenever $\alpha > 1$ and $\Delta_\infty$ satisfies*
$$\alpha < \frac{2n\lambda}{\Delta_\infty} + 1, \text{ and } (10.9\alpha^2 - 1.8\alpha - 9.1) < \frac{4n\lambda}{\Delta_\infty^2}, \quad (3)$$
*Algorithm 1 with noise parameter $\lambda$ satisfies $(\alpha, \tau)$-RDP with $\tau = \frac{1.2\alpha+1}{2} \cdot \frac{c}{2n\lambda}$.*

We extend Theorem 5 to the multi-dimensional setting using Lemma 1.

COROLLARY 1. *Suppose that each participant's data point $x_i$ is d-dimensional and satisfies*
$$\sum_{j=1}^{d} \left( |x_{i,j}|^2 + (|x_{i,j}| - \lfloor |x_{i,j}| \rfloor) - (|x_{i,j}| - \lfloor |x_{i,j}| \rfloor)^2 \right) \leq c, \quad (4)$$
*and $\|\lceil |x_i| \rceil\|_\infty \leq \Delta_\infty$. Then, whenever $\alpha > 1$ and $\Delta_\infty$ satisfies Eq. (3), Algorithm 2 with noise parameter $\lambda$ satisfies $(\alpha, \tau)$-RDP with $\tau = \frac{1.2\alpha+1}{2} \cdot \frac{c}{2n\lambda}$.*

In practice, the constraints in Eq. (4) and $\|\lceil |x_i| \rceil\|_\infty \leq \Delta_\infty$ can be enforced by clipping, as we explain in Section 4. The maximum value of the $\mathcal{L}_\infty$ clipping bound $\Delta_\infty$ is computed from Eq. (3). Next, we present the utility guarantee incurred by Algorithm 2, which follows from Corollary 1.

COROLLARY 2. *Suppose that each participant's data point $x_i$ is d-dimensional and satisfies Eq. (4), $\|\lceil |x_i| \rceil\|_\infty \leq \Delta_\infty$, $\alpha > 1$, and $\Delta_\infty$ satisfies Eq. (3). Then, when satisfying $(\alpha, \tau)$-RDP, the error incurred by Algorithm 1 is*
$$Err_\mathcal{M} = \frac{1.2\alpha+1}{2} \cdot \frac{dc}{\tau} + \sum_{i=1}^{n}\sum_{j=1}^{d} \left( |x_{i,j}| - \lfloor |x_{i,j}| \rfloor - (|x_{i,j}| - \lfloor |x_{i,j}| \rfloor)^2 \right).$$

We briefly comment on Corollary 2. We define $p_{i,j} := |x_{i,j}| - \lfloor |x_{i,j}| \rfloor$, which is the probability of increasing the absolute value $|x_{i,j}|$ by 1 for the $i$-th participant. Then, the overall error incurred by dSMM is:
$$Err_\mathcal{M} = \frac{(1.2\alpha+1) \cdot dc}{2\tau} + \sum_{i=1}^{n}\sum_{j=1}^{d} (p_{i,j} - p_{i,j}^2).$$

The first term of $Err_\mathcal{M}$ can be viewed as the error due to enforcing differential privacy. Note that the leading multiplier $(1.2\alpha + 1)/2$ is only slightly larger (i.e., by a constant factor) than of the approach injecting continuous Gaussian noise, which is $\alpha/2$. The second error term is the overall variance of all the Bernoulli trials performed on the participant side. This error term can be seen as the integer approximation error, which exists even without enforcing differential privacy.

---

**Algorithm 3:** Federated learning with Skellam mixture mechanism

**Input:** Private dataset of training records $X = (x_1, \ldots, x_n)$; initial model parameters $\theta$; secure aggregation protocol $\mathcal{A}$.

**Parameters:** Sampling parameter $q$; number of iterations $T$; noise parameter $\lambda$; scale parameter $\gamma$; clipping thresholds $c$ and $\Delta_\infty$; modulus $m \in \mathbb{N}$.

1 **for** $h \in 1 \ldots T$ **do**
2     The server shares the current model parameters $\theta$ to all participants.
3     $B \xleftarrow{u.a.r} \{1, 2, \ldots, n\}$.    // sample a subset of participants uniformly at random from all participants using Poisson sampling with rate $q$
4     **for** $i \in B$ **do**
5        $g_i \leftarrow \nabla_\theta(r_i)$.    // gradient computation
6        $z_i \leftarrow$ Algorithm 4($g_i$).    // SMM on the participant side
7     $\bar{z} \leftarrow \mathcal{A}(\{z_i\}_{i \in B})$.    // secure aggregation
8     $\bar{g}^* \leftarrow$ Algorithm 6($\bar{z}$).    // gradient sum retrieval by the server
9     $\theta \leftarrow Update(\theta, \bar{g}^*)$.    // model update based on the approximate gradient sum

**Output:** $\theta$ model parameters learnt on $X$.

---

## 4 FEDERATED LEARNING WITH SKELLAM MIXTURE MECHANISM

In this section, we apply our Skellam mixture mechanism (SMM) to enforce DP on federated learning with distributed SGD. We assume that the participants have access to a black-box secure aggregation protocol, following the convention in [4, 31]. The training process is outlined in Algorithm 3. In each iteration, the server releases the current model parameters to all participants (Line 2 in Algorithm 3). Then, a random subset of participants, whose identities are not known to the server, is selected (Line 3). Each participant in the selected subset then computes the gradients based on the current model weights and her own data (Line 5), and invoke Algorithm 4 for gradient perturbation (Line 6). After that, the secure aggregation protocol computes the sum of the perturbed gradients (Line 7) of the randomly selected participants. Finally, the server retrieves the perturbed gradient sum and updates the model (Lines 8 and 9). We omit additional details on the updating process (*e.g.*, learning rate schedule, weight decay) as they do not affect the general framework or the privacy guarantees. After repeating the above process for $T$ iterations, the training terminates, and the server obtains the final model weights $\theta$.

In what follows, we explain the participant procedure for perturbing gradients (Algorithm 4) and the server procedure for reconstructing the perturbed gradient sum (Algorithm 6). Each participant $i$ first randomly rotates the private vector using a Walsh-Hadamard matrix [29] and a public random sign vector $\xi$ shared among all participants and the server (Line 1 in Algorithm 4), which is also used in previous solutions [3, 4, 31]. Each dimension of the resulting gradient follows a Sub-Gaussian distribution with variance

**Algorithm 4:** participant procedure for perturbing gradients

**Input:** Private gradient $g_i \in \mathbb{R}^d$
**Parameters:** Noise parameter $\lambda$; scale parameter $\gamma$; clipping thresholds $c$ and $\Delta_\infty$; modulus $m \in \mathbb{N}$.
**Public randomness:** Uniformly random sign vector
$$\xi \in \{-1, +1\}^d.$$

1. $g_i \leftarrow H_d D_\xi g_i$.   // random rotation, where $H \in \{-1/\sqrt{d}, +1/\sqrt{d}\}^{d\times d}$ is a Walsh-Hadamard matrix satisfying $H^T H = I$ and $D_\xi \in \{-1, 0, +1\}^{d \times d}$ is a diagonal matrix with $\xi$ on the diagonal
2. $g_i \leftarrow \gamma \cdot g_i$.   // scaling
3. $g_i \leftarrow clip(g_i)$.   // clip $g_i$ as in Algorithm 5
4. **for** $j \in 1 \ldots d$ **do**
5. $\quad p_{i,j} = g_{i,j} - \lfloor g_{i,j} \rfloor$.
6. $\quad$ Sample $y_{i,j}$ from a Bernoulli trial with success probability $p_{i,k}$.
7. $\quad$ **if** $y_{i,j} = 0$ **then**
8. $\quad\quad g^*_{i,j} \leftarrow \lfloor g_{i,j} \rfloor + Sk(\lambda, \lambda)$.
9. $\quad$ **else**
10. $\quad\quad g^*_{i,j} \leftarrow \lfloor g_{i,j} \rfloor + 1 + Sk(\lambda, \lambda)$.
11. $\quad z_{i,j} \leftarrow g^*_{i,j} \mod m$.

**Output:** $z_i \in \mathbb{Z}_m^d$ for the secure aggregation protocol.

---

**Algorithm 5:** participant procedure for clipping gradients

**Input:** Private gradient $g_i \in \mathbb{R}^d$
**Parameters:** Clipping thresholds $c$ and $\Delta_\infty$.

1. $v_i \leftarrow 0$. // initialize the helper vector for clipping.
2. **for** $j \in 1..d$ **do**
3. $\quad v_{i,j} = \frac{g_{i,j}}{|g_{i,j}|} \cdot \left(|g_{i,j}|^2 + |g_{i,j}| - \lfloor |g_{i,j}| \rfloor + (|g_{i,j}| - \lfloor |g_{i,j}| \rfloor)^2\right)$.
   // map $g_i$ to $v_i$.
4. $v_i \leftarrow \min(1, \frac{c}{\|v_i\|_1}) \cdot v_i$.   // $\mathcal{L}_1$ clip and re-scale
5. **for** $j \in 1 \ldots d$ **do**
6. $\quad g'_{i,j} = \lfloor \sqrt{|v_{i,k}|} \rfloor$.   // compute the integer part
7. $\quad p'_{i,j} = \frac{y}{2g'_{i,j}+1}$   // compute the fraction part
8. $\quad g_{i,j} \leftarrow \frac{v_{i,j}}{|v_{i,j}|} \cdot (g'_{i,j} + p'_{i,j})$. // compose two parts
9. **for** $j \in 1 \ldots d$ **do**
10. $\quad g_{i,j} \leftarrow \frac{g_{i,j}}{|g_{i,j}|} \cdot \min(\Delta_\infty, |g_{i,j}|)$.   // $\mathcal{L}_\infty$ clip.

**Output:** $g_i$ the clipped gradient.

---

**Algorithm 6:** Server procedure of estimating gradient sum

**Input:** Private vector $\bar{z} = (\sum_{i \in B} z_i \mod m) \in \mathbb{Z}_m^d$ via secure aggreagtion
**Parameters:** Noise parameter $\lambda$; scale parameter $\gamma$; clipping thresholds $c$ and $\Delta_\infty$; modulus $m \in \mathbb{N}$.
**Public randomness:** Uniformly random sign vector
$$\xi \in \{-1, +1\}^d.$$

1. Map $\bar{z} \in \mathbb{Z}_m^d$ to $\bar{z}' \in [-m/2, m/2]^d \cap \mathbb{Z}^d$.
2. $\bar{g}^* \leftarrow \frac{1}{\gamma} \cdot D_\xi H_d^T \bar{z}'$.

**Output:** $\bar{g}^*$ the estimated gradient sum.

---

$O(\|g_i\|_2^2 / d)$, where $g_i$ is the participant's private gradient value. Specifically, each dimension is concentrated around 0 when $d$ is large, e.g., tens of thousand for neural networks. Essentially, this operation flattens the gradient and limits the probability of overflowing when computing the sum of gradients. We refer the reader to [4, 31] for detailed discussions.

After that, the participant scales the rotated vector and clips the scaled vector (Lines 2 and 3). We will explain the clipping procedure shortly (outlined in Algorithm 5). Then, for each $k$-th coordinate in the clipped vector, the participant samples one bit from the Bernoulli distribution of success probability $g_{i,k} - \lfloor g_{i,k} \rfloor$, where $g_{i,k}$ is the $k$-th element of the rotated vector $g_i$ (Lines 5 and 6 in Algorithm 4). If the Bernoulli trial fails, the participant samples a noise following the Skellam distribution $Sk(\lambda, \lambda)$ and shift the outcome to $\lfloor g_{i,k} \rfloor$ (Lines 7 and 8); otherwise, the participant shifts the same outcome to $\lceil g_{i,k} \rceil$ (Lines 9 and 10). Finally, the participant applies element-wise modulo operation on the noisy vector (Line 11). Essentially, this step restricts the output from the participant to $\mathbb{Z}_m^d$, and enforces a $\log_2 m$-bit communication constraint per dimension, both of which are required by the secure aggregation protocol.

Next, the participants collectively compute the sum of their output noisy vectors through a secure aggregation protocol, and reveal the sum $(\sum_{i=1}^n z_i \mod m)$ to the server. As we have mentioned, parameter $m$ can be seen as the per dimension communication for secure aggregation protocol. Although a larger $m$ helps preserve information on the noisy gradients, such an $m$ increases the communication cost, slowing down the aggregation process (especially with a communication-intensive secure aggregation protocol) as well as the model training overall. The problem is exacerbated when the participant is a mobile device with metered Internet connection. Hence, in practice it is often preferable to set a relatively small $m$, e.g., $2^8$ in our experiments, which is equivalent to a communication constraint of one-byte per dimension.

After obtaining the sum $(\sum_{i=1}^n z_i \mod m)$, the server first unwraps the modulo operation (Line 1 in Algorithm 6). In particular, values in $\{m/2, m/2 + 1, \ldots, m - 1\}$ are mapped back to $\{-m/2, -m/2 + 1, \ldots, -1\}$, respectively; and values in $\{0, 1, \ldots, m/2 - 1\}$ remain unchanged. This is because in Line 11 in Algorithm 4, values in $\{-m/2, -m/2+1, \ldots, -1\}$ are mapped to $\{m/2, m/2+1, \ldots, m-1\}$, respectively; and values in $\{0, 1, \ldots, m/2-1\}$ are mapped to themselves. We refer the reader to [31] for a detailed discussion on this issue. Then, the server reverses the rotation and scaling performed on the participant side (Line 2 in Algorithm 6), obtaining an unbiased estimate for the gradient sum.

Next, we explain the clipping procedure outlined in Algorithm 5. Recall that clipping is a standard step introduced in DPSGD [2] to bound the sensitivity of deep learning with DP. The clipping procedure in this work is slightly different from that in DPSGD, which clips the $\mathcal{L}_2$ norm of the gradient. The difference is due to the different privacy guarantee of SMM (see Theorem 5 and Corollary 1). Recall that the privacy guarantee of $d$-dimensional SMM relies on the following property of the input data $g_i$:

$$\lceil |g_i| \rceil \leq \Delta_\infty, \text{ and}$$

$$\sum_{j=1}^d \left( |g_{i,j}|^2 + (|g_{i,j}| - \lfloor |g_{i,j}| \rfloor) - (|g_{i,j}| - \lfloor |g_{i,j}| \rfloor)^2 \right) \leq c.$$

Accordingly, this requires a different clipping procedure. The first property is easy to enforce. For example, for $\Delta_\infty = 1$ and $x_i =$

−1.9, we simply increase the $x_i$ to −1. The second property is more complicated to enforce, as we explained next. For each participant, we first construct a helper vector $v_i$. In particular, each dimension of $v_i$ is computed as follows:

$$v_{i,j} = \frac{g_{i,j}}{|g_{i,j}|} \cdot \left( |g_{i,j}|^2 + |g_{i,j}| - \lfloor |g_{i,j}| \rfloor + (|g_{i,j}| - \lfloor |g_{i,j}| \rfloor)^2 \right),$$

for $j = 1 \ldots, d$ (Line 3 in Algorithm 5). For completeness, we define $\frac{0}{0} = 1$. Next, we clip vector $v_i$ based on its $\mathcal{L}_1$ norm in the standard way (Line 4 in Algorithm 5). Finally, we re-map the clipped vector to its original form (Lines 5 to 8 in Algorithm 5) and clip each dimension of the vector by $\Delta_\infty$ (Line 10 in Algorithm 5).

### 4.1 Privacy Analysis

In this section, we analyze the privacy guarantee of Algorithm 3. Observe that each iteration of Algorithm 3 can be seen as running the Skellam mixture mechanism on a random subset of gradients. This is because none of the model sharing (Line 2 in Algorithm 3), gradient sum reconstruction (Line 8 in Algorithm 3), or model updating (Line 9 in Algorithm 3) procedures incurs any additional privacy loss, as the updated model can be reconstructed by the constructed perturbed gradient sum, which, in turn, can be computed from the perturbed gradient sum released from the secure aggregation protocol. In addition, since the identities of the random subset of participants are not known to the server, the privacy guarantee benefits from amplification by subsampling. (We refer the reader to [31] for a detailed discussion on this issue.) Hence, the privacy guarantee of Algorithm 3 follows by applying the composition (Lemma 1) and the amplification (Lemma 2) results to the privacy analysis of SMM (Corollary 1). A formal statement of the privacy guarantees of Algorithm 3 is as follows.

THEOREM 6 (PRIVACY GUARANTEE OF ALGORITHM 3). *For sampling parameter $q$; sampled subset $B$; number of iterations $T$; noise parameter $\lambda$; and clipping thresholds $c$ and $\Delta_\infty$, for any $\alpha > 1$ and $\Delta_\infty$ satisfies*

$$\alpha < \frac{2|B|\lambda}{\Delta_\infty} + 1, \text{ and } (10.9\alpha^2 - 1.8\alpha - 9.1) < \frac{4|B|\lambda}{\Delta_\infty^2}, \quad (5)$$

*Algorithm 3 satisfies $(\alpha, \tau)$-RDP with*

$$\tau = T \cdot \frac{1}{\alpha - 1} \cdot$$
$$\log \left( (1-q)^{\alpha-1}(\alpha q - q - 1) + \sum_{l=2}^{\alpha} \binom{\alpha}{l}(1-q)^{\alpha-l}q^l e^{(l-1)\tau(l)} \right),$$

*where $\tau(l)$ is defined as $\tau(l) := \frac{1.2l+1}{2} \cdot \frac{c}{2|B|\lambda}$, for $l = 2, \ldots, \alpha$.*

## 5 RELATED WORK

As mentioned in Section 1, existing work on federated learning with differential privacy has mostly considered the non-MPC settings where real-value noise can be used. To our knowledge, there are only four prior studies [3, 4, 31, 34] on using integer noise to achieve DP in federated learning. In what follows, we revisit the solutions in [3, 4, 31, 34], and compare them with our SMM.

**cpSGD [4].** cpSGD lets each participant inject binomial noise (*i.e.*, the sum of multiple binary values drawn from independent Bernoulli trials) to her discretized gradients to satisfy DP. Similar to Gaussian noise in the continuous domain, binomial noise can also be easily aggregated, as the sum of multiple i.i.d. binomial values also follows a binomial distribution. This property simplifies the privacy reasoning for cpSGD, as it allows us to focus on the aggregated binomial noise in the sum of all participants' gradients, without the need to analyze each participant's binomial noise separately. However, the privacy analysis of cpSGD in [4] is based on $(\epsilon, \delta)$-DP instead of RDP, which leads to relatively loose privacy bounds for federated learning, since it is difficult to derive the exact $(\epsilon, \delta)$-DP guarantee of an iterative algorithm with subsampling (e.g., SGD).

Another limitation of cpSGD is that it assumes the input to to be integer-valued. For any non-integer input $x$, the method requires a *stochastic rounding* [4] of $x$, which often leads to a considerable increase in sensitivity. For example, if $x = \{0.01, 0.01, \ldots, 0.01\} \in \mathbb{R}^d$, then each dimension of $x$ is rounded 1 with 0.01 probability, and to 0 with 0.99 probability. This approach ensures that each rounded value's expectation equals the original value, but the rounded values could be significantly larger than the original ones. In particular, in the worst case when each dimension of $x$ is rounded to 1, the $\mathcal{L}_2$ norm of $x$ is increased from $0.01 \cdot \sqrt{d}$ to $\sqrt{d}$ after the rounding. In other words, even if each participant's gradient vector has an $\mathcal{L}_2$ norm at most $0.01 \cdot \sqrt{d}$, the sum of all participant's rounded gradients could have an $\mathcal{L}_2$ sensitivity of $\sqrt{d}$. This significantly increases the amount of noise required by cpSGD to achieve differential privacy, resulting in an unfavorable trade-off between privacy and utility.

**Distributed Discrete Gaussian (DDG) mechanism [31].** To mitigate the limitations of cpSGD, Kairouz *et al.* [31] propose DDG, a method that utilizes *discrete Gaussian distributions* [12] instead of binomial distributions for noise generation. In particular, a discrete Gaussian distribution has a similar PDF to a continuous Gaussian distribution, but is defined over the integer domain. The main advantage of using discrete Gaussian noise is that it can achieve RDP, which makes it much easier to derive a tight privacy bound of DDG for iterative algorithms with subsampling.

Similar to cpSGD, DDG also assumes that the inputs are integer-valued, and, hence, requires stochastic rounding of non-integers. To alleviate the sensitivity increase incurred by rounding, DDG applies a *conditional rounding* approach as follows. First, given an input $x \in \mathbb{R}^d$ with bounded $\mathcal{L}_2$ norm $\Delta_2$ (otherwise DDG clips the input) and the scale parameter $\gamma$, DDG scales the input $x$ and obtains $\gamma x$. After that, DDG performs a stochastic rounding on $\gamma x$. If the rounded version of the scaled input has an $\mathcal{L}_2$ norm larger than

$$\sqrt{\gamma^2 \Delta_2^2 + d/4 + \sqrt{2\log(1/\beta)}(\gamma \Delta_2 + \sqrt{d}/2)}, \quad (6)$$

for some fixed $\beta$ (explained soon), then DDG discards it and re-generates another stochastically rounded version. The procedure is repeated until the above requirement is met. The hyperparameter $\beta$ ranging from 0 to 1 controls the trade-off between bias and sensitivity increase in the conditional rounding process. To see this, note that the expectation of the rounded value is generally not equal to the original value (since rounded values failing the above condition in Eq. (6) are rejected), which adversely affects the accuracy of the output of DDG. A smaller $\beta$ leads to a lower bias but higher sensitivity increase, which, in turn, leads to a higher amount of noise needed to satisfy DP, and vice versa. This conditional rounding approach ensures that the rounding operation does not incur a significant increase of the $\mathcal{L}_2$ sensitivity, but the increase is still $O(\sqrt{d})$. In addition, the conditional rounding operation introduces a hyperparameter $\beta$, which is difficult to tune under DP. The authors of [31] recommend fixing $\beta$ to $e^{-0.5}$, which is done in our experiments.

**Skellam Mechanism [3].** In Ref. [3], Agarwal *et al.* propose to sample noise from a Skellam distribution instead of a discrete Gaussian distribution. Since the sum of independent Skellam noises still follows Skellam distribution (see Section 2.1), the privacy reasoning of distributed Skellam noise is straightforward, unlike DDG. In particular, the paper shows that adding Skellam-distributed noise to integers also achieves RDP. However, for non-integer inputs, the Skellam mechanism in [3] still requires the conditional rounding approach introduced in [31]. Consequently, its accuracy also suffers from the sensitivity increase, as well as the bias introduced by conditional rounding.

**Comparisons with SMM.** Compared to the aforementioned methods, one major advantage of SMM is that it does not rely on an additional stochastic rounding [4] or conditional rounding [3, 31] step to handle non-integer inputs. Instead, SMM directly takes any $x \in \mathbb{R}^d$ as input, and outputs a noisy version $x^*$ of $x$ whose expectation equals $x$, without incurring a significant increase in sensitivity. Accordingly, SMM injects a smaller amount of noise while achieving the same level of privacy as its competitors. As a consequence, SMM is able to obtain much more accurate results than cpSGD [4], DDG [31], and the Skellam mechanism [3], especially in settings where communication is constrained to low bitwidths. In particular, in such situations, the quantization granularity is set to a coarse level (i.e., a small scale parameter $\gamma$) to avoid overflow. Such a coarse quantization granularity leads to a relatively large sensitivity increase compared to the quantized gradients. For this reason, the perturbation noise in cpSGD, DDG, and Skellam due to rounding is rather high in such low-bitwidth settings, resulting in much lower model utility than SMM. We validate this claim with experiments in the next section.

In addition, compared with DPSGD [2], SMM involves only one additional hyper parameter: the scale parameter $\gamma$, which controls the trade-off between communication cost and utility. Note that this trade-off does not exist in DPSGD as it is a solution for the centralized setting. Once $\gamma$ is determined, we can compute the clipping threshold $c$ for SMM as $c = \gamma^2 \cdot \Delta_2^2$ for some constant $\Delta_2$, which corresponds to setting the $\mathcal{L}_2$ clipping norm to $\Delta_2$ in DPSGD [2]. In addition, the $\mathcal{L}_\infty$ clipping bound $\Delta_\infty$ for SMM is computed from Eq. (5). In contrast, both DDG [31] and the Skellam mechanism [3] include an additional hyperparameter $\beta$. In these algorithms, parameter $\beta$ controls the trade-off between bias and sensitivity in their conditional rounding process, as mentioned earlier. A poor choice of $\beta$ may adversely impact the performance of these algorithms; meanwhile, hyperparameter tuning is rather challenging under the differential privacy requirement.

We also note that our theoretical analysis of SMM is substantially different from that in [3], due to the inherent differences between the Skellam mixture distribution used in SMM and the Skellam distribution used in [3]. In addition, even for the special case of integer inputs, the privacy guarantee of SMM (see Theorems 3 and 4) differs from that of the Skellam mechanism in [3], because we use different proof techniques from those in [3]. While the techniques used in [3] is also non-trivial, our result for integer inputs is cleaner, and is of independent interest, as we have mentioned in Section 3.3.

## 6 EXPERIMENTS

We evaluate the performance of SMM on the distributed sum estimation problem and two basic machine learning tasks. For simplicity, all experiments are done using the approximate samplers for Discrete Gaussian and Skellam from the TensorFlow libraries, which are based on floating point approximations. Compared with exact samplers, approximate samplers are faster. We include a detailed discussion on this issue in Appendix A in the technical report version [7].

### 6.1 Distributed Sum Estimation

As a simple application, we first evaluate the performance of solution SMM on the distributed sum estimation problem described in Section 3.1, given a private $d$-dimensional input dataset. Following the experiment setting in [31], we generate a synthetic dataset containing $n = 100$ data points uniformly sampled from a $d$-dimensional $\mathcal{L}_2$ sphere. We set the dimension to $d = 65536$, and the radius to $r = 1$ (namely, the $\mathcal{L}_2$ sensitivity of input is 1). The participants release their noisy sum under distributed DP. We report the mean squared error (mse) over all dimensions. Our evaluation uses the $(\epsilon, \delta)$-DP (Definition 2) definition instead of RDP (Definition 3, since $(\epsilon, \delta)$-DP is a classic definition of differential privacy, and a competitor cpSGD supports the former but not the latter. We fix $\delta$ to $10^{-5}$, and vary the privacy parameter $\epsilon$ from $\{1, 2, 3, 4, 5\}$. For DDG, Skellam, and SMM, we first compute the privacy guarantee using RDP, and then convert the guarantee to $(\epsilon, \delta)$-DP using Lemma 3 (the optimal RDP order is chosen from integers from 2 to 100).

For DDG, Skellam, cpSGD, and SMM, we vary the communication bitwidth per dimension from $\{10, 12, 14, 16, 18\}$. Correspondingly, $m$ varies from $\{2^{10}, 2^{12}, 2^{14}, 2^{16}, 2^{18}\}$ (see line 11 in Algorithm 4). For $m$ equals to $2^{10}, 2^{12}, 2^{14}, 2^{16}$, and $2^{18}$, we vary the scale parameter $\gamma$ in $\{4, 8\}$, $\{16, 32\}$, $\{64, 128\}$, $\{256, 512\}$, and $\{1024, 2048\}$, respectively (see line 2 in Algorithm 4). For SMM, the clipping threshold $c$ is set to $\gamma^2 r^2$ with $r = 1$. Additionally, we compute the $\mathcal{L}_\infty$ clipping bound for SMM using Eq. (3), based on the optimal RDP order. For Skellam and DDG, the $\mathcal{L}_2$ clipping bound is set to $\Delta_2 = \sqrt{\gamma^2 r^2 + d/4 + \sqrt{2\log(1/\beta)}(\gamma r + \sqrt{d}/2)}$, with $r = 1$, $d = 65536$, and $\beta = \exp(-0.5)$, as suggested in [31]. In terms of $\mathcal{L}_1$ clipping bound for Skellam and DDG, we have that $\Delta_1 \leq \min\left(\sqrt{d} \cdot \Delta_2, \Delta_2^2\right)$, following [31]. Note that we do not perform an actual $\mathcal{L}_1$ clipping step for the rounded gradients, as the above relationship between $\mathcal{L}_2$ and $\mathcal{L}_1$ norms automatically holds for all integer-valued vectors. Similarly, for cpSGD, the $\mathcal{L}_1$ norm is bounded by $\sqrt{d}$ times the $\mathcal{L}_2$ norm, following its original implementation. We also include continuous Gaussian, which is a solution for the centralized DP setting, as a strong baseline.

The results are shown in Figure 1. When the communication bitwidth is limited (i.e., when $m = 2^{10}, 2^{12}, 2^{14}$), SMM significantly outperforms all its competitors, as demonstrated in Figures 1 (a), (b), (c), (f), (g), and (h). When $m = 2^{16}$ and $\gamma = 256$, SMM achieves comparable performance as DDG and Skellam, as shown in Figure 1 (d). When both $m$ and $\gamma$ are large, SMM performs slightly worse than DDG and Skellam, which obtain almost the same accuracy as the strong baseline continuous Gaussian, as we see from Figures 1 (i), (e), and (j). Finally, Skellam and DDG has similar performance under all settings, and cpSGD incurs rather high error ($> 10^4$), and falls outside the error range shown in the figures. Below, we briefly explain the reasons for the above results.

As mentioned in Section 5, existing solutions for distributed DP incur high sensitivity overhead due to stochastic rounding (in cpSGD) or conditional rounding (in DDG and Skellam). To be more specific, the sensitivity overhead is roughly 1 per dimension, which is non-negligible compared to the scaled data, especially when the data dimension is large (e.g., $d = 65536$) and when the quantization granularity is coarse (i.e., small $\gamma$) under small bitwidths (i.e., small $m$). This sensitivity increase leads to stronger perturbations for cpSGD, DDG, and Skellam, and explains why SMM performs the best in settings with small bitwidths. As the bitwidth increases with $\gamma$, the above-mentioned sensitivity overhead becomes negligible compared with the scaled data. As a result, Skellam and DDG yield almost the same accuracy as continuous Gaussian. In the meantime, SMM performs slightly worse than continuous Gaussian, DDG, and Skellam. This is because SMM always incurs a slightly larger error than continuous Gaussian, according to Corollary 2, where there is an extra factor of 1.2 leading the error term of SMM.

## 6.2 Federated Learning

Next, we evaluate the performance of the proposed solution SMM on FL with DP (Algorithm 3) on two classic benchmark datasets: MNIST [35] and Fashion MNIST [54], which contain grayscale images of handwritten digits and clothing, respectively. Both datasets represent 10-class classification tasks with 60,000 training data records. We regard each data record in the training data as a participant. Our evaluation uses the $(\epsilon, \delta)$-DP, as we have explained earlier. We fix $\delta$ to $10^{-5}$, and vary the privacy parameter $\epsilon$ from $\{1, 2, 3, 4, 5\}$. In particular, for cpSGD, we apply both linear composition and advanced composition [22] for privacy accounting and choose the stronger guarantee between them. We have also included the strong central-model DPSGD [2] as a baseline.

For both MNIST [35] and Fashion MNIST [54], we train a three-layer neural network with fully connected layers and ReLu activation, following previous work [4]. We set the number of neurons per layers to 80, resulting in a model with $d = 63,610$ weights. For DDG, Skellam, cpSGD, and SMM, we vary the communication constraint $m$ from $\{2^6, 2^8, 2^{10}\}$, where $m = 2^8$ corresponds to one byte per parameter. For each $m$, we vary the scaling parameter $\gamma$ in $\{m/32, m/16, m/8, m/4, m/2, m\}$ (see line 2 in Algorithm 4). For cpSGD, DDG, Skellam, and the centralized algorithm DPSGD, we use the same $\mathcal{L}_2$ clipping norm of 1 for the original real-valued gradients. For the scaled gradients in DDG and Skellam, we set $\mathcal{L}_2$ clipping bound to $\sqrt{\gamma^2 \Delta_2^2 + d/4 + \sqrt{2\log(1/\beta)}(\gamma \Delta_2 + \sqrt{d}/2)}$, with $\Delta_2 = 1$, $d = 65536$, and $\beta = \exp(-0.5)$. For SMM, we set the clipping threshold $c$ to $\gamma^2 \Delta_2^2$, with $\Delta_2 = 1$, similar to its competitors. In terms of the $\mathcal{L}_\infty$ clipping bound for SMM, we compute $\Delta_\infty$ from Eq. (3) using the optimal order of $\alpha$. We also vary batch size $|B|$ from $\{120, 240, 480, 960\}$. The model is trained for 4 epochs, i.e., when $|B|$ equals to 120, 240, 480, and 960, we train the model for 2000, 1000, 500, and 250 rounds, respectively. For all experiments, we use the Adam optimizer [33] with learning rate $\eta = 0.005$. We do not tune the hyper-parameters in favor of any particular solution and omit additional experiments on hyper parameter tuning, e.g., model structure, learning rate, clipping norm, optimizer, training epochs, etc. We remark that our approach is compatible with existing differentially private parameter tuning techniques [28, 36, 45], which is an orthogonal topic to this paper. We report the average test accuracy over 5 runs. The results are shown in Figures 2 and 3.

Overall, the results are consistent with the those for distributed sum estimation, and lead to similar conclusions as before, i.e., SMM has a clear performance advantage over its competitors with small bitwidths, and the performance gap gradually closes as the bitwidth increases.

Specifically, when $m = 2^6$, SMM is the only method that achieves meaningful accuracy under all settings of privacy parameter $\epsilon$, batch size $|B|$, and scale ratio $\gamma$ (see Figures 2(a), (b), and (c), and Figures 3(a), (b), and (c)). This is because the scale of the noise injected in DDG, Skellam, and cpSGD is so large that it causes floating point number overflow, destroying the utility of the resulting gradient sum.

When $m = 2^8$ (i.e., one byte per parameter), SMM also achieves significantly higher accuracy compared to its competitors. In particular, in Figures 2(d) and 3(d), we fix the scale parameter to $\gamma = 64$ and the batch size to $|B| = 240$. When $\epsilon = 1$, DDG and Skellam yield very low utility due to floating point number overflows, while SMM achieves much higher utility that is close to that of DPSGD (i.e., the gap is less than 10%). As $\epsilon$ increases (indicating weaker privacy protection), the performance gap between SMM and its competitors becomes less dramatic. This is because with a higher $\epsilon$, the required noise scale for the competitors becomes smaller, to the point that it no longer causes floating point number overflows. Nevertheless, there remains a noticable performance gap, since the noise scale of SMM is still significantly lower than that of its competitors. In particular, when $\epsilon = 3$, the accuracy improvement of SMM over DDG and Skellam is around 6% and 10% for MNIST and Fashion MNIST, respectively, while the accuracy gap between SMM and the centralized baseline DPSGD is only around 3%.

The performance gap between SMM and the centralized DPSGD algorithm exists, even when $\epsilon$ reaches as high as 5. Not that at this point, the noise required to satisfy DP no longer dominates the total amount of perturbations; instead, the relatively coarse quantization granularity (i.e., caused by a small $\gamma$) becomes a significant factor. As we demonstrate shortly, this accuracy gap gradually closes with a larger bitwidth and/or a large scale ratio $\gamma$.

In Figures 2(e) and 3(e), we fix the privacy parameter $\epsilon = 3$ and the scale parameter $\gamma = 64$, and vary the batch size $|B|$ from $\{120, 240, 480, 960\}$. SMM is the only algorithm that consistently achieves comparable accuracy with DPSGD under all settings of $|B|$. In particular, when $|B| = 960$, the accuracy improvement of SMM over DDG and Skellam is around 30% and 20% for MNIST and

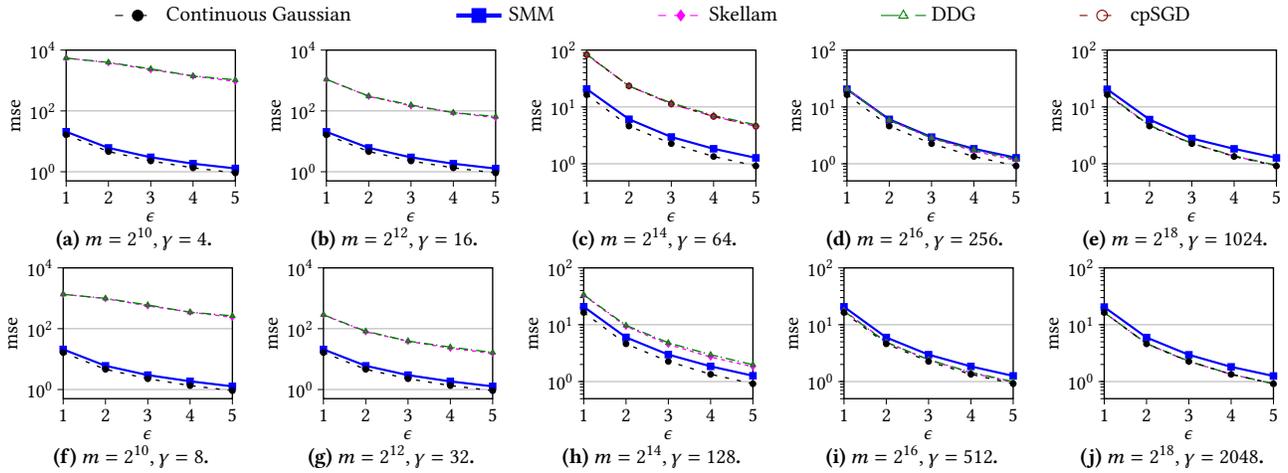

Figure 1: Evaluations on synthetic data with varying privacy parameter $\epsilon$, scale parameter $\gamma$, and communication constraint $m$.

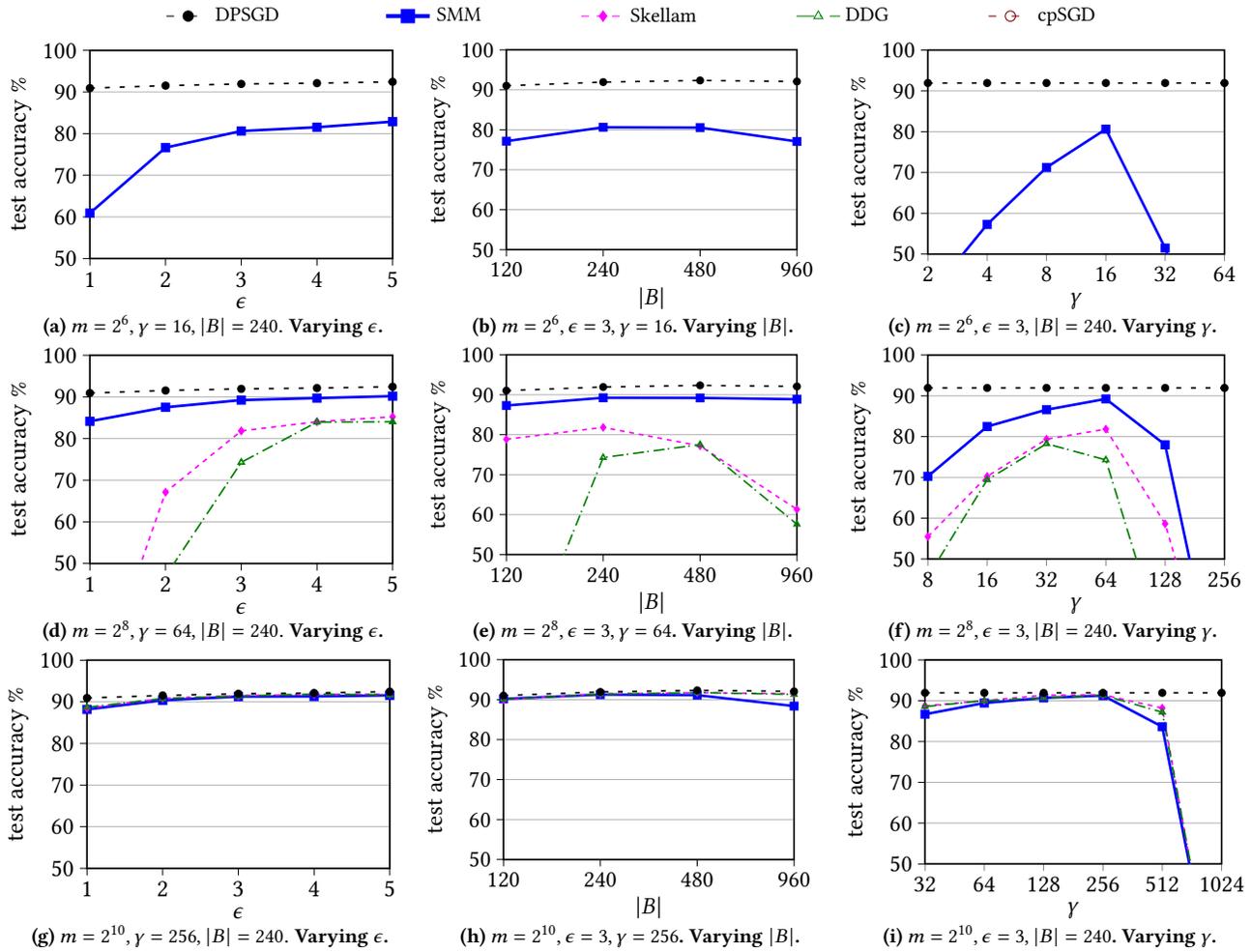

Figure 2: Evaluations on MNIST with varying communication constraint $m$, privacy parameter $\epsilon$, scale parameter $\gamma$, and batch size $|B|$.

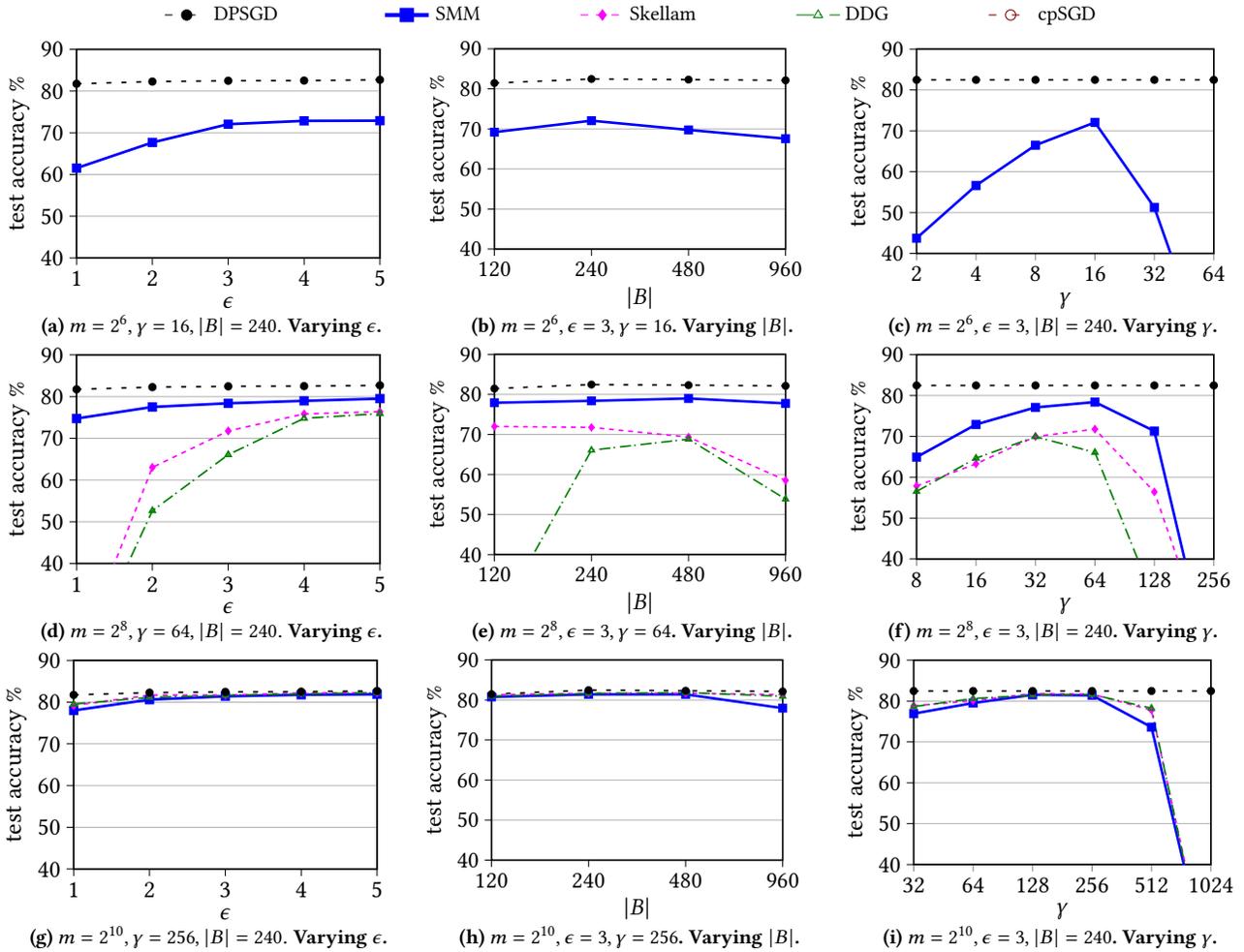

Figure 3: Evaluations on Fashion MNIST with varying communication constraint $m$, privacy parameter $\epsilon$, scale parameter $\gamma$, and batch size $|B|$.

Fashion MNIST, respectively. Finally, we fix the privacy parameter and the batch size, and vary the scale parameter $\gamma$ in Figures 2(f) and 3(f). The results show consistent accuracy improvement with varying $\gamma$. We also note that as $\gamma$ increases from 8 to 256, the accuracy of SMM first increases then decreases. On the one hand, as $\gamma$ increases, the gradient weights becomes more fine-grained and contains more information, leading to higher accuracy; on the other hand, as $\gamma$ increases, a larger amount of noise is required to satisfy DP. When $\gamma = 256$, the noisy gradient weights exceed the one-byte communication constraint, causing utility degradation. The same performance pattern can be observed for Skellam and DDG.

When the communication bitwidth is sufficiently large (e.g., when $m = 2^{10}$), we observe that while DDG and Skellam achieve almost the same accuracy as DPSGD, there is a small accuracy gap between SMM and DPSGD. For example, as we see in Figures 2(g) and 3(g), there is a 0.5 and 1 percent accuracy gap for MNIST and Fashion MNIST between SMM and DPSGD for $\epsilon \geq 2$. In addition, there are also noticeable accuracy gaps when $|B| = 960$ and when $\gamma = 512$ in Figures 2(h) and (i) and 3(h) and (i). Overall, as the bitwidth increases, the performance increase of DDG and Skellam is much more significant than SMM, whose performance is relatively stable with different bitwidths. Lastly, under all settings, the accuracy of cpSGD is rather low (< 20%), and falls outside the accuracy range shown in the figures.

## 7 CONCLUSION

This paper presents the Skellam mixture mechanism (SMM), a novel solution for enforcing differential privacy on machine learning models built through an MPC-based federated learning process using distributed stochastic gradient descent. Compared to existing solutions, SMM achieves composable and scalable privacy guarantee without increasing the sensitivity of input. Extensive experiments, performed on both a synthetic dataset and two classic benchmark datasets, as well as various practical settings, demonstrate the consistent and significant accuracy gains SMM over existing solutions under restrictive communication constraints.

For future work, we plan to further reduce the constant factor in the privacy analysis for SMM to improve model utility under the same level of privacy protection. Another promising direction is to open up the black box of the MPC protocol and perform careful privacy analysis with considerations for the details of the MPC protocol, which might help lower the noise level further, leading to a more favorable privacy-utility trade-off for federated learning.

## ACKNOWLEDGMENTS

This work was supported by the Ministry of Education, Singapore (Number MOE2018-T2-2-091), A*STAR, Singapore (Number A19E3b0099), and Qatar National Research Fund Qatar Foundation (Number NPRP11C-1229-170007). Any opinions, findings and conclusions or recommendations expressed in this material are those of the authors and do not reflect the views of the funding agencies.

## Appendix A    EXACT SAMPLER FOR POISSON

We present the exact sampler for Poisson noise with a rational parameter $\lambda$, where

$$\lambda := m_x/m_y, \text{ with } m_x, m_y \in \mathbb{Z}, m_x \geq 0, m_y > 0.$$

Following Ref. [12], we adopt the convention that $RandInt(n)$, which uniformly samples an integer from $\{1, \ldots, n\}$, is the only accessible randomness for the sampler. There are two notable boundary cases of input $\lambda$. The first is when $\lambda = 1$, in which case an exact sampler for Poisson(1) is given by Duchon and Duvignau in [20], outlined in Algorithm 7. The second case is when $\lambda < 1$, presented in Algorithm 8, whose idea comes from the fact that $Poisson(\lambda)$ ($\lambda < 1$) follows the same distribution as the sum of $N$ Bernoulli variates of success probability $\lambda$, with $N$ following a Poisson distribution of $\lambda = 1$ (see page 487 of [19]). Hence, when $\lambda < 1$, we can reduce the sampling process to first call Algorithm 7 and then call the Bernoulli sampler of success probability $m_x/m_y$ multiple times, depending on the outcome of Algorithm 7. Note that a Bernoulli sampler of success probability $p_x/p_y$ returns 1 if $RandInt(p_y) <= p_x$, and 0 otherwise, as shown in Algorithm 9.

We use Algorithm 7 and Algorithm 8 as building blocks for the general case sampler, outlined in Algorithm 10. The overall idea of Algorithm 10 is as follows. If $m_x = 0$ (i.e., $\lambda = 0$), we simply return 0 (Line 3). Otherwise, while $m_x > m_y$ (i.e., $\lambda > 1$), we repeatedly reduce $\lambda$ by 1 and add $Poisson(1)$ to the returned value, until $\lambda < 1$ (Lines 5 and 6). This is because the sum of Poisson variates is a Poisson. Finally, we reach the case when $m_x < m_y$ (i.e., $0 \leq \lambda < 1$), and we simply call Algorithm 8 with parameter $m_x/m_y$ and add the outcome to the return value (Lines 8 and 9) if $m_x > 0$ (i.e., $\lambda > 0$).

### A.1    Running Time for Sampling Noise

Next, we empirically compare the running time of the exact Skellam sampler and exact discrete Gaussian sampler, as well as their respective approximate versions. We use Algorithm 10 to construct exact Skellam variates, since a Skellam sample can be obtained as the difference between two Poisson samples. In our Python implementation, we adopt the convention that **randrange()**, which uniformly samples an integer from the specified range, is the only accessible randomness, following the implementation of Ref. [32].

For the non-exact samplers based on floating point approximations, we use the implementations from the TensorFlow libraries. We generate $10^5$ samples and report the time of generating all samples, averaged over 10 runs. For the exact samplers, the samples are generated sequentially. For the non-exact samplers, the entire array of $10^5$ samples are generated at once as a tensor. The experiments are conducted on a Linux machine with four Intel Xeon Gold 6240 CPUs and 376 gigabytes of memory.

We vary the variance for Skellam and Discrete Gaussian distributions from $\{16, 8, 4, 2, 1\}$. This setting covers the particularly interesting parameter regime considered in this paper where SMM achieves better privacy-utility trade-off than DGM (i.e., when the noise variance is small and the bitwidth is small). The results are shown in Table 1. We observe that the exact sampler for Skellam is more efficient than the exact sampler for Discrete Gaussian, especially when the variance of the noise distribution is small. Meanwhile, the approximate sampler for Skellam is also much faster than the approximate sampler for Discrete Gaussian. The results suggests that it is more efficient to use Skellam than Discrete Gaussian for both practical DP applications (use the exact version) and DP illustrations (use the approximate version) under small bitwidth. Finally, the approximate samplers for both distributions are much faster than the exact ones. Further improving the running time for the exact samplers is an interesting future work direction.

---

**Algorithm 7:** Exact sampler for Poisson(1)

**Output:** $k$, a sample from Poisson(1).
1  $n \leftarrow 1, g \leftarrow 0, k \leftarrow 1$.
2  **while** *true* **do**
3      $i \leftarrow RandInt(n+1)$. // uniformly sample from $\{1, \ldots, n+1\}$
4      **if** $i = n+1$ **then**
5         $k \leftarrow k + 1$.
6      **else if** $i > g$ **then**
7         $k \leftarrow k - 1, g \leftarrow n + 1$.
8      **else**
9         return $k$.
10     $n \leftarrow n + 1$.

---

**Algorithm 8:** Exact sampler for Poisson($\lambda$), $0 < \lambda < 1$

**Input:** $m_x, m_y \in \mathbb{Z}, m_y > 0, 0 < m_x < m_y$.
**Output:** $k$, a sample from Poisson($\lambda$), with $\lambda := m_x/m_y$.
1  $k \leftarrow 0$.
2  $n \leftarrow Poisson(1)$. // Sample $n$ from Poisson(1) using Algorithm 7.
3  **for** $i = 1..n$ **do**
4      $k \leftarrow k + Bernoulli(m_x/m_y)$.
5  return k.

## Appendix B    DISCRETE GAUSSIAN MIXTURE MECHANISM

So far, our discussion focuses on injecting random noise to the gradients following the Skellam mixture distribution. The insightful reader might have found that the algorithmic framework of our

**Algorithm 9:** Exact sampler for Bernoulli($p$), $0 \leq p \leq 1$

**Input:** $p_x, p_y \in \mathbb{Z}, p_y > 0, 0 \leq p_x \leq p_y$.
**Output:** A sample from Bernoulli($p$), with $p := p_x/p_y$.

1. $n \leftarrow RandInt(p_y)$. // Uniformly sample from $\{1, 2, \ldots, p_y\}$
2. **if** $n \leq p_x$ **then**
3.     return 1.
4. **else**
5.     return 0.

---

**Algorithm 10:** Exact sampler for general Poisson($\lambda$), $\lambda \geq 0$.

**Input:** $m_x, m_y \in \mathbb{Z}, m_x \geq 0, m_y > 0$.
**Output:** $k$, a sample from Poisson($\lambda$), with $\lambda := m_x/m_y$.

1. $k \leftarrow 0$.
2. **if** $m_x = 0$ **then**
3.     return $k$.
4. **while** $m_x \geq m_y$ **do**
5.     $n \leftarrow Poisson(1)$. // Sample from Poisson(1) using Algorithm 7
6.     $k \leftarrow k + n, m_x \leftarrow m_x - m_y$.
7. **if** $m_x > 0$ **then**
8.     $n \leftarrow Poisson(m_x/m_y)$. // Sample from Poisson($m_x/m_y$) using Algorithm 8
9.     $k \leftarrow k + n$.
10. return $k$

---

**Algorithm 11:** One-dimensional Discrete Gaussian mixture mechanism (1DGM)

**Input:** A set of private values $\{x_1, \ldots, x_n \mid x_i \in \mathbb{R}\}$.
**Parameters:** Noise parameter $\lambda$.

1. **for** $i \in 1..n$ **do**
2.     $p_i = x_i - \lfloor x_i \rfloor$.
3.     Sample $y_i$ from a Bernoulli trial with success probability $p_i$.
4.     **if** $y_i = 0$ **then**
5.         $x_i^* \leftarrow \lfloor x_i \rfloor + \mathcal{N}_\mathbb{Z}(\mathbf{0}, \sigma^2)$.
6.     **else**
7.         $x_i^* \leftarrow \lfloor x_i \rfloor + 1 + \mathcal{N}_\mathbb{Z}(\mathbf{0}, \sigma^2)$.
8. $\bar{x}^* \leftarrow SecAgg((x_1^*, \ldots, x_n^*))$.
**Output:** $\bar{x}^*$.

---

proposed method is not limited to the Skellam distribution. In this appendix, we adapt the proposed mechanism to work with the Discrete Gaussian noise [12], leading to a new method that we call the Discrete Gaussian Mixture (DGM).

Algorithms 11 and 12 present the pseudo code for the 1D and multi-dimensional versions of DGM, respectively, which closely resemble their counterparts in SMM, i.e., Algorithms 1 and 2. The only difference between DGM and SMM lies in the noise distribution. In particular, in Lines 5 and 7 of Algorithm 11 and in Lines 6 and 8 of Algorithm 12, we inject Discrete Gaussian noise instead of symmetric Skellam noise. Next, we present the privacy guarantees of Algorithm 11 and Algorithm 12.

---

**Algorithm 12:** Multi-dimensional Discrete Gaussian mixture mechanism (dDGM)

**Input:** A set of private values $\{x_1, \ldots, x_n \mid x_i \in \mathbb{R}^d\}$.
**Parameters:** Noise parameter $\lambda$, data dimension $d$.

1. **for** $i \in 1..n$ **do**
2.     **for** $j \in 1..d$ **do**
3.         $p_{i,d} = x_{i,d} - \lfloor x_{i,d} \rfloor$.
4.         Sample $y_{i,j}$ from a Bernoulli trial with success probability $p_{i,j}$.
5.         **if** $y_{i,j} = 0$ **then**
6.             $x_{i,d}^* \leftarrow \lfloor x_{i,d} \rfloor + \mathcal{N}_\mathbb{Z}(\mathbf{0}, \sigma^2)$.
7.         **else**
8.             $x_{i,d}^* \leftarrow \lfloor x_{i,d} \rfloor + 1 + \mathcal{N}_\mathbb{Z}(\mathbf{0}, \sigma^2)$.
9. $\bar{x}^* \leftarrow SecAgg((x_1^*, \ldots, x_n^*))$.
**Output:** $\bar{x}^*$.

---

| Variance | 32 | 16 | 8 | 4 | 2 | 1 |
|---|---|---|---|---|---|---|
| Exact Skellam | 26.77 | 15.09 | 9.32 | 6.41 | 4.98 | 5.38 |
| Exact DG | 16.98 | 17.66 | 18.17 | 19.68 | 19.69 | 24.52 |
| TF Skellam | 0.004 | 0.004 | 0.004 | 0.004 | 0.003 | 0.003 |
| TF DG | 0.053 | 0.053 | 0.053 | 0.052 | 0.052 | 0.052 |

**Table 1:** Average running time (in seconds) of generating $10^5$ samples for Skellam and Discrete Gaussian (DG) using exact and non-exact samplers (from TensorFlow libraries).

## B.1 Privacy Analysis

We first briefly review the privacy guarantee provided by distributed Discrete Gaussian noise [31]. The main difference between using Skellam noise and Discrete Gaussian noise in the distributed setting is that the sum of independent Skellam noises is still Skellam, whereas the sum of independent Discrete Gaussian noises no longer follows the Discrete Gaussian distribution. Formally, we define $Z_{n,\sigma^2}$ to be the sum of $n$ independent Discrete Gaussian noises of zero mean and variance $\sigma^2$. Formally,

$$Z_{n,\sigma^2} := \sum_{i=1}^n Z_i, \text{ where } Z_i \overset{\text{i.i.d.}}{\sim} \mathcal{N}_\mathbb{Z}(0, \sigma^2).$$

where $\mathcal{N}_\mathbb{Z}(0, \sigma^2)$ denotes the Discrete Gaussian distribution with mean zero and variance $\sigma^2$. Then, the distribution of $Z_{n,\sigma^2}$ is close to, but not exactly the same as, the distribution of $Z_{1,n\sigma^2}$. Canonne et al. [12] derive their difference as

$$\tau_n := 10 \sum_{k=1}^{n-1} \exp\left(-2\pi^2 \sigma^2 \frac{k}{k+1}\right), \tag{7}$$

and we have the following theorem [31].

**Theorem 7.** For any integer $s \in \mathbb{Z}$, and any $\alpha > 1$, we have

$$D_\alpha(s + Z_{n,\sigma^2} \| Z_{n,\sigma^2}) \leq \min\left\{\frac{\alpha s^2}{2n\sigma^2} + \tau_n, \frac{\alpha}{2}\left(\frac{s}{\sqrt{n}\sigma} + \tau_n\right)^2\right\}.$$

In a nutshell, Theorem 7 states that the sum of $n$ independent Discrete Gaussian noises sampled from $\mathcal{N}_\mathbb{Z}(0, \sigma^2)$ provides approximately the same privacy guarantee as a single Discrete Gaussian noise sampled from $\mathcal{N}_{\mathbb{Z}}(0, n\sigma^2)$, which is $\alpha s^2/(2 \cdot n\sigma^2)$ in terms of Rényi divergence, with approximation error $\tau_n$. Theorem 7 is the foundation of our privacy analysis for Algorithm 11, presented as follows.

THEOREM 8. *Suppose that each client's data point $x_i$ satisfies*

$$|x_i|^2 + (|x_i| - \lfloor|x_i|\rfloor) - (|x_i| - \lfloor|x_i|\rfloor)^2 \leq c,$$

*and $\lceil|x_i|\rceil \leq \Delta_\infty$. Then, whenever $\alpha > 1$, and $\Delta_\infty$ satisfies*

$$\frac{\alpha \Delta_\infty^2}{2n\sigma^2} + \tau_n < \frac{0.1}{\alpha - 1}, \text{ and } \left(\frac{\Delta_\infty}{\sqrt{n}\sigma} + \tau_n\right)^2 < \frac{0.2}{\alpha^2 - \alpha}, \quad (8)$$

*Algorithm 11 with noise parameter $\sigma$ satisfies $(\alpha, \tau)$-RDP with*

$$\tau = \min\left\{\frac{1.1\alpha c}{2n\sigma^2} + 1.1\tau_n, \frac{1.1\alpha c}{2n\sigma^2} + \frac{1.1\alpha\Delta_\infty}{\sqrt{n}\sigma}\tau_n + 1.1\tau_n^2\right\},$$

The proof of Theorem 8 follows the same logic as the proof of Theorem 5; their difference lies in the privacy guarantees provided by the sum of Discrete Gaussian (see Theorem 7) and Skellam (see Theorem 3) noises. A proof sketch for Theorem 8 is presented in Appendix C.7. Next, we extend Theorem 8 to the multi-dimensional setting using Lemma 1.

COROLLARY 3. *Suppose that each client's data point $x_i$ is $d$-dimensional and satisfies*

$$\sum_{j=1}^{d}\left(|x_{i,j}|^2 + (|x_{i,j}| - \lfloor|x_{i,j}|\rfloor) - (|x_{i,j}| - \lfloor|x_{i,j}|\rfloor)^2\right) \leq c,$$

$\|x_i\|_1 \leq \Delta_1$, *and $\lceil|x_i|\rceil \leq \Delta_\infty$. Then, whenever $\alpha > 1$, and $\Delta_\infty$ satisfies Eq. (8), Algorithm 12 with noise parameter $\sigma$ satisfies $(\alpha, \tau)$-RDP with*

$$\tau = \min\left\{\frac{1.1\alpha c}{2n\sigma^2} + 1.1d\tau_n, \frac{1.1\alpha c}{2n\sigma^2} + \frac{1.1\alpha\Delta_1}{\sqrt{n}\sigma}\tau_n + 1.1d\tau_n^2\right\},$$

### B.2 FL with Discrete Gaussian Mixture

Next, we apply DGM to FL with distributed SGD. Overall, the training procedure is similar to FL with SMM; in the following, we briefly explain the main modifications compared to SMM. The overall procedure is outlined as in Algorithm 13, and the client's procedure for noise injection is outlined as in Algorithm 14. In particular, in Line 6 of Algorithm 13, we replace the perturbation algorithm SMM with DGM on the client side. In Lines 8 and 10 of Algorithm 14, the clients inject discrete Gaussian noise instead of Skellam. The server procedure and the clipping procedure remains the same as in SMM. We present the privacy guarantee of Algorithm 13.

THEOREM 9 (PRIVACY GUARANTEE OF ALGORITHM 13). *For sampling parameter $q$, sampled subset $B$, number of iterations $T$, noise parameter $\sigma$, clipping thresholds $c$, $\Delta_1$ and $\Delta_\infty$, for any $\alpha$ satisfying $\alpha > 1$, $1.1\alpha c/(2|B|\sigma^2) < 0.1 - 1.1d\tau_{|B|}$, and $1.1 \cdot \alpha c/(2 \cdot |B|\sigma^2) + 1.1 \cdot \alpha\Delta_\infty\tau_{|B|}/(\sqrt{|B|}\sigma) < 0.1 - 1.1 \cdot d\tau_{|B|}^2$, Algorithm 13 satisfies $(\alpha, \tau)$-RDP with*

$$\tau = T \cdot \frac{1}{\alpha - 1} \cdot$$

$$\log\left((1-q)^{\alpha-1}(\alpha q - q - 1) + \sum_{l=2}^{\alpha}\binom{\alpha}{l}(1-q)^{\alpha-l}q^l e^{(l-1)\tau(l)}\right),$$

---

**Algorithm 13:** Federated learning with Discrete Gaussian mixture

**Input:** Private dataset of training records $R = (r_1, \ldots, r_n)$; initial model parameters $\theta$; secure aggregation protocol $\mathcal{A}$.

**Parameters:** Sampling parameter $q$; number of iterations $T$; noise parameter $\sigma$; scale parameter $\gamma$; clipping thresholds $c$ and $\Delta_\infty$; modulus $m \in \mathbb{N}$.

1 **for** $h \in 1 \ldots T$ **do**
2      The server shares the current model parameters $\theta$ to all clients.
3      $B \xleftarrow{u.a.r} \{1, 2, \ldots, n\}$.    // sample a subset of clients uniformly at random from all clients using Poisson sampling with rate $q$
4      **for** $i \in B$ **do**
5          $g_i \leftarrow \nabla_\theta(r_i)$.    // gradient computation
6          $z_i \leftarrow$ Algorithm 14$(g_i)$.    // DGM on the client side
7      $\bar{z} \leftarrow \mathcal{A}(\{z_i\}_{i \in B})$.    // secure aggregation
8      $\bar{g}^* \leftarrow$ Algorithm 6$(\bar{z})$. // gradient sum retrieval by the server
9      $\theta \leftarrow Update(\theta, \bar{g}^*)$.    // model update based on the approximate gradient sum

**Output:** $\theta$ model parameters learnt on $X$.

---

**Algorithm 14:** Client procedure for perturbing gradients (DGM)

**Input:** Private gradient $g_i \in \mathbb{R}^d$
**Parameters:** Noise parameter $\sigma$; scale parameter $\gamma$; clipping thresholds $c$ and $\Delta_\infty$; modulus $m \in \mathbb{N}$.
**Public randomness:** Uniformly random sign vector $\xi \in \{-1, +1\}^d$.

1 $g_i \leftarrow H_d D_\xi g_i$.    // random rotation, where $H \in \{-1/\sqrt{d}, +1/\sqrt{d}\}^{d \times d}$ is a Walsh-Hadamard matrix satisfying $H^T H = I$ and $D_\xi \in \{-1, 0, +1\}^{d \times d}$ is a diagonal matrix with $\xi$ on the diagonal
2 $g_i \leftarrow \gamma \cdot g_i$.    // scaling
3 $g_i \leftarrow clip(g_i)$.    // clip $g_i$ as in Algorithm 5
4 **for** $k \in 1 \ldots d$ **do**
5      $p_{i,k} = g_{i,k} - \lfloor g_{i,k} \rfloor$.
6      Sample $y_{i,k}$ from a Bernoulli trial with success probability $p_{i,k}$.
7      **if** $y_{i,k} = 0$ **then**
8          $g^*_{i,k} \leftarrow \lfloor g_{i,k} \rfloor + \mathcal{N}_{\mathbb{Z}}(0, \sigma^2)$.
9      **else**
10          $g^*_{i,k} \leftarrow \lfloor g_{i,k} \rfloor + 1 + \mathcal{N}_{\mathbb{Z}}(0, \sigma^2)$.
11      $z_{i,k} \leftarrow g^*_{i,k} \mod m$.

**Output:** $z_i \in \mathbb{Z}_m^d$ for the secure aggregation protocol.

---

where for $l = 2, \ldots, \alpha$, $\tau(l)$ is defined as

$$\tau(l) = \min\left\{\frac{1.1lc}{2|B|\sigma^2} + 1.1d\tau_{|B|}, \frac{1.1lc}{2|B|\sigma^2} + \frac{1.1l\Delta_1}{\sqrt{|B|}\sigma}\tau_{|B|} + 1.1d\tau_{|B|}^2\right\}.$$

### B.3 Experiments

We empirically evaluate the performance of DGM for distributed sum estimation and FL tasks. For all experiments, we enforce $(\epsilon, \delta)$-DP, as explained in Section 6. We fix $\delta$ to $10^{-5}$ and vary $\epsilon$ in $\{1, 2, 3, 4, 5\}$. For distributed sum estimation, we vary communication constraint $m$ in $2^{10}, 2^{14}, 2^{18}$ and scale ratio $\gamma$ in 4, 64, 1024,

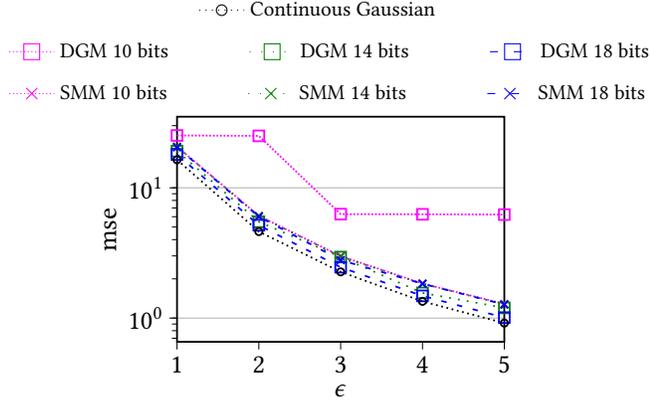
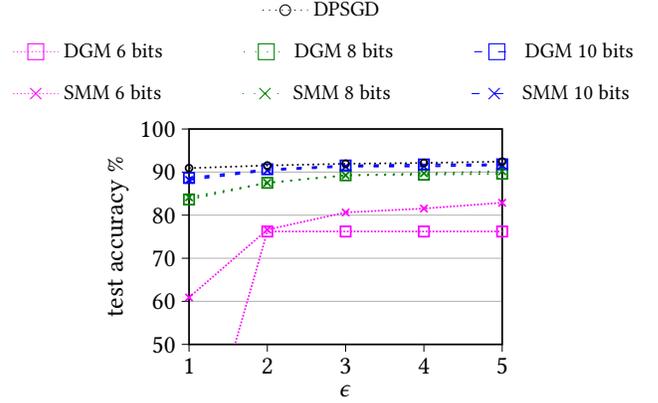

Figure 4: Evaluations distributed sum estimation using synthetic dataset with varying communication constraint $m$, privacy parameter $\epsilon$, scale parameter $\gamma$.

(a) MNIST

(b) Fashion MNIST

Figure 5: Evaluations on FL tasks with Fashion MNIST and MNIST with varying communication constraint $m$, privacy parameter $\epsilon$, scale parameter $\gamma$.

respectively. For FL, we consider two basic tasks MNIST and Fashion MNIST. The model setup is the same as in Section 6. To be more specific, we fix the batch size to 240 and train the model for 1000 rounds (i.e., 4 epochs) using the Adam optimizer with learning rate 0.005. We vary the communication constraint per dimension from {6, 8, 10} bits, and the scale parameter $\gamma$ from {16, 64, 256}, respectively. Similar to SMM, the clipping threshold $c$ is set to $\gamma^2$ for DGM, corresponding to $\Delta_2 = 1$ in DPSGD. The $\mathcal{L}_\infty$ clipping bound is computed from Eq. (3). In terms of the bound for $\mathcal{L}_1$ norm of input, we consider the general relationship $\Delta_1 \le \sqrt{d}\Delta_2$, where $\Delta_2^2 \le c = \gamma^2$. Hence, we have $\Delta_1 \le \sqrt{d}\gamma$.

Overall, the results of DGM are comparable to those of SMM, except for small bandwidths. One reason is that the noise parameter $\sigma$ for DGM is integer-valued in the current implementation in the TensorFlow Privacy library[1]. For example, if the $\sigma$ is computed as 0.9 based on privacy constraints, then $\sigma$ is rounded up to its nearest integer, 1, for the actual perturbation. Hence, for small bandwidth and small scale parameter, $\sigma$ may remain constant with different $\epsilon$'s (e.g., $\epsilon = 1, 2$ and $\epsilon = 3, 4, 5$ under 10-bit bandwidth in Figure 4, and $\epsilon = 2, 3, 4, 5$ under 6-bit bandwidth in Figure 5). Another reason is that the sum of discrete Gaussian noises is no longer a discrete Gaussian [31], and the divergence between the sum of discrete Gaussian noises and a single discrete Gaussian noise is large especially when the noise parameter $\sigma$ is small, which can happen under the small bandwidth settings. As a result, the sum of discrete Gaussian noises are not as concentrated as Skellam, leading to worse performance than SMM and even overflows under strong privacy settings ($\epsilon = 1$ under 6-bit bandwidth in Figure 5).

## Appendix C PROOFS
### C.1 Proof of Theorem 4
We recall and prove Theorem 3, which states the Rényi divergence between two Skellam distributions.

---
[1]https://github.com/tensorflow/privacy/blob/master/tensorflow_privacy/privacy/dp_query/distributed_discrete_gaussian_query.py

THEOREM (RÉNYI DIVERGENCE OF SKELLAM DISTRIBUTIONS). *For any integer $s \in \mathbb{Z}$ satisfying $|s| \le \Delta_\infty$, and any $1 < \alpha < 2\lambda/\Delta_\infty + 1$, we have*

$$D_\alpha(s + Sk(\lambda, \lambda) \| Sk(\lambda, \lambda)) \le \frac{1.09\alpha + 0.91}{2} \cdot \frac{s^2}{2\lambda}.$$

PROOF. Without loss of generality we consider $s > 0$, since the Skellam distribution $Sk(\lambda, \lambda)$ is symmetric with respect to the origin. We let $Z \sim Sk(\lambda, \lambda)$ and compute $\Phi = \exp((\alpha - 1)D_\alpha(Z + s \| Z))$. We first present several useful inequalities, whose proofs can be found in Section D. Let

$$Q_{s,t}(v) := \prod_{i=1}^{s} \frac{v + i + \sqrt{(v+i)^2 + t^2}}{t}. \qquad (9)$$

According to Lemma 6 in Appendix D, we have:

$$\frac{I_{|v|}(t)}{I_{|v+s|}(t)} \le Q_{s,t}(v).$$

In addition, for any $0 < a \le w$, according to Lemmata 7-9 in Appendix D, we have

$$\frac{w + \sqrt{w^2 + 1}}{(a - w) + \sqrt{(a-w)^2 + 1}} \le e^{2w-a}, \qquad (10)$$

and $\left(w + \sqrt{w^2 + 1}\right) \cdot \left((a - w) + \sqrt{(a-w)^2 + 1}\right) \le e^a. \qquad (11)$

By the definition of Skellam distributions and Eq. (9)

$$\Phi = \sum_{z=-\infty}^{\infty} \left(\frac{I_{|z|}(2\lambda)}{I_{|z+s|}(2\lambda)}\right)^{\alpha-1} \Pr[Z=z]$$

$$\leq \sum_{z=-\infty}^{\infty} (Q_{s,2\lambda}(z))^{\alpha-1} \Pr[Z=z]$$

$$= \sum_{|z| \leq s} (Q_{s,2\lambda}(z))^{\alpha-1} \Pr[Z=z]$$

$$+ \sum_{z>s} \left((Q_{s,2\lambda}(z))^{\alpha-1} + (Q_{s,2\lambda}(-z))^{\alpha-1}\right) \Pr[Z=z].$$

We compute

$$(Q_{s,2\lambda}(z))^{\alpha-1} + (Q_{s,2\lambda}(-z))^{\alpha-1}$$
$$= (Q_{s,2\lambda}(z)Q_{s,2\lambda}(-z))^{\frac{\alpha-1}{2}} \cdot \left(R_{s,2\lambda}(z) + \frac{1}{R_{s,2\lambda}(z)}\right),$$

where

$$R_{s,2\lambda}(z) := \left(\frac{Q_{s,2\lambda}(z)}{Q_{s,2\lambda}(-z)}\right)^{\frac{\alpha-1}{2}}.$$

By Eq. (10), when $z \geq i$,

$$\frac{z+i+\sqrt{(z+i)^2+(2\lambda)^2}}{-z+i+\sqrt{(-z+i)^2+(2\lambda)^2}} = \frac{\frac{z+i}{2\lambda}+\sqrt{(\frac{z+i}{2\lambda})^2+1}}{\frac{-z+i}{2\lambda}+\sqrt{(\frac{-z+i}{2\lambda})^2+1}} < \exp(\frac{z}{\lambda}).$$

Therefore, when $z \geq s$, we have

$$\frac{Q_{s,2\lambda}(z)}{Q_{s,2\lambda}(-z)} < \exp\left(\frac{sz}{\lambda}\right) \text{ and } R_{s,2\lambda}(z) < \exp\left(\frac{s(\alpha-1)z}{2\lambda}\right).$$

Since the function $v \mapsto v + \frac{1}{v}$ is increasing,

$$R_{s,2\lambda}(z) + \frac{1}{R_{s,2\lambda}(z)} < \exp\left(\frac{s(\alpha-1)z}{2\lambda}\right) + \exp\left(-\frac{s(\alpha-1)z}{2\lambda}\right).$$

By Eq. (11), when $z \geq i$,

$$\frac{z+i+\sqrt{(z+i)^2+(2\lambda)^2}}{2\lambda} \cdot \frac{-z+i+\sqrt{(-z+i)^2+(2\lambda)^2}}{2\lambda}$$

$$= \left(\frac{z+i}{2\lambda}+\sqrt{\left(\frac{z+i}{2\lambda}\right)^2+1}\right) \cdot \left(\frac{-z+i}{2\lambda}+\sqrt{\left(\frac{-z+i}{2\lambda}\right)^2+1}\right)$$

$$\leq \exp\left(\frac{i}{\lambda}\right).$$

Hence, when $z \geq s$, we have $Q_{s,2\lambda}(z) \cdot Q_{s,2\lambda}(-z) < \exp\left(\frac{s(s+1)}{2\lambda}\right)$. Meanwhile, when $-s < z < s$, according to Lemma 10 in the appendix, we have $Q_{s,2\lambda}(z) \leq \exp\left(\frac{s(s+z)}{2\lambda}\right)$. Therefore,

$$\Phi \leq \exp\left(\frac{s(s+1)(\alpha-1)}{4\lambda}\right).$$

$$\sum_{z=s}^{\infty} \left(\exp(\frac{s(\alpha-1)z}{2\lambda}) + \exp(-\frac{s(\alpha-1)z}{2\lambda})\right) \Pr[Z=z]$$

$$+ \sum_{z=-s+1}^{s-1} \exp\left(\frac{s(\alpha-1)(s+z)}{2\lambda}\right) \Pr[Z=z]$$

$$\leq \exp\left(\frac{s^2(\alpha-1)}{2\lambda}\right) \sum_{z=-\infty}^{\infty} \exp\left(\frac{s(\alpha-1)z}{2\lambda}\right) \Pr[Z=z]$$

$$= \exp\left(\frac{s^2(\alpha-1)}{2\lambda}\right) \mathbb{E}\left[\exp\left(\frac{s(\alpha-1)Z}{2\lambda}\right)\right].$$

The moment generating function of $Sk(\lambda, \lambda)$ is $\mathbb{E}[e^{tZ}] = e^{\lambda(e^t+e^{-t}-2)}$. When $0 < t < 1$, we have $e^t + e^{-t} - 2 < 1.09t^2$. Thus, when $\lambda > \frac{s(\alpha-1)}{2}$,

$$\mathbb{E}\left[\exp\left(\frac{s(\alpha-1)Z}{2\lambda}\right)\right] \leq \exp\left(1.09\lambda(\frac{s(\alpha-1)}{2\lambda})^2\right) = \exp\left(\frac{1.09s^2(\alpha-1)^2}{4\lambda}\right).$$

As a result, when $\lambda > \frac{s(\alpha-1)}{2}$, we have

$$\Phi \leq \exp\left(\frac{s^2(\alpha-1)}{2\lambda} + \frac{1.09s^2(\alpha-1)^2}{4\lambda}\right). \qquad \square$$

### C.2 Proof of Theorem 5

We sketch the proof of Theorem 5. Missing details can be found in Appendices C.3 and C.4. We start by considering the Rényi divergence $D_\alpha(\text{1SMM}(X) \parallel \text{1SMM}(X'))$ for any two neighboring datasets $X$ and $X'$. The following lemma shows that this Rényi divergence only depends on the tuple that differs in the two neighboring datasets.

LEMMA 4. *Let $X = \{x_1, \ldots, x_n\}$ and $X' = \{x_1, \ldots, x_n, x_{n+1}\}$ be any two neighboring datasets, where each $x_i \in \mathbb{R}$. If $\alpha > 1$, then,*

$$D_\alpha(\text{1SMM}(X) \parallel \text{1SMM}(X'))$$
$$\leq D_\alpha(\text{1SMM}(\{0, \ldots, 0\}) \parallel \text{1SMM}(\{0, \ldots, 0, x_{n+1}\})), \text{ and}$$
$$D_\alpha(\text{1SMM}(X') \parallel \text{1SMM}(X))$$
$$\leq D_\alpha(\text{1SMM}(\{0, \ldots, 0, x_{n+1}\}) \parallel \text{1SMM}(\{0, \ldots, 0\})). \quad \square$$

By Lemma 4, to establish the privacy guarantee of Algorithm 1, we only need to derive the maximum values of $A_\alpha$ and $B_\alpha$ defined as follows:

$$A_\alpha = D_\alpha(\text{1SMM}(\{0, \ldots, 0\}) \parallel \text{1SMM}(\{0, \ldots, 0, x_{n+1}\})), \text{ and}$$
$$B_\alpha = D_\alpha(\text{1SMM}(\{0, \ldots, 0, x_{n+1}\}) \parallel \text{1SMM}(\{0, \ldots, 0\})).$$

Namely, it suffices to consider the dataset containing $n$ zeros and its neighboring dataset containing $n$ zeros and an extra tuple $x_{n+1}$. Without loss of generality, we assume $x_{n+1} > 0$, and denote $D_\alpha(s + Sk(\lambda, \lambda) \parallel Sk(\lambda, \lambda))$ as $D_\alpha^\lambda(s)$ for integer $s \in \mathbb{Z}$. The following lemma establishes upper bounds of $A_\alpha$ and $B_\alpha$.

LEMMA 5. *For $\alpha > 1$,*

$$A_\alpha \leq (1 - x_{n+1} + \lfloor x_{n+1} \rfloor) \cdot D_\alpha^\lambda(\lfloor x_{n+1} \rfloor)$$
$$+ (x_{n+1} - \lfloor x_{n+1} \rfloor) \cdot D_\alpha^\lambda(\lceil x_{n+1} \rceil), \text{ and}$$

$$B_\alpha \leq \frac{1}{\alpha-1} \cdot \ln \Big((1 - x_{n+1} + \lfloor x_{n+1} \rfloor) \cdot \exp\big((\alpha-1) \cdot D_\alpha^\lambda(\lfloor x_{n+1} \rfloor)\big)$$
$$+ (x_{n+1} - \lfloor x_{n+1} \rfloor) \cdot \exp\big((\alpha-1) \cdot D_\alpha^\lambda(\lceil x_{n+1} \rceil)\big)\Big). \quad \square$$

It remains to bound $D_\alpha^\lambda(s)$, i.e., $D_\alpha(s + Sk(\lambda, \lambda) \parallel Sk(\lambda, \lambda))$ for any integer $s \in \mathbb{Z}$, which is computed in Theorem 3. Based on Theorem 3 and Lemmata 4 and 5, we can compute the privacy guarantee of Algorithm 1, which we recall as follows.

THEOREM. *Suppose that each client's data point $x_i$ satisfies*

$$|x_i|^2 + (|x_i| - \lfloor |x_i| \rfloor) - (|x_i| - \lfloor |x_i| \rfloor)^2 \leq c$$

*and $\lceil |x_i| \rceil \leq \Delta_\infty$. Then, whenever $\alpha$ satisfies*

$$1 < \alpha < 2n\lambda/\Delta_\infty + 1, \text{ and } (10.9\alpha^2 - 1.8\alpha - 9.1) < 4n\lambda/\Delta_\infty^2,$$

*Algorithm 1 with noise parameter $\lambda$ satisfies $(\alpha, \tau)$-RDP with $\tau = \frac{1.2\alpha+1}{2} \cdot \frac{c}{2n\lambda}$.*

PROOF. By Theorem 3 and Lemmata 4 and 5, we can prove the theorem by showing that for any $x_{n+1} > 0$ satisfying the constraints above, the following inequalities hold:

$$A_\alpha \leq \frac{1.2\alpha + 1}{2} \cdot \frac{c}{2n\lambda}, \text{ and } B_\alpha \leq \frac{1.2\alpha + 1}{2} \cdot \frac{c}{2n\lambda},$$

where $A_\alpha$ and $B_\alpha$ are defined as in Lemma 5. Let $p = x_{n+1} - \lfloor x_{n+1} \rfloor$. Then, the first inequality follows from the fact that

$$(1-p) \cdot (\lfloor x_{n+1} \rfloor)^2 + p \cdot (\lfloor x_{n+1} \rfloor + 1)^2 = x_{n+1}^2 + p - p^2,$$

and $1.09\alpha + 0.91 < 1.1\alpha \leq 1.2\alpha$ for $\alpha > 1$.

To prove the second inequality, we compute

$$\Xi = \exp((\alpha - 1)B_\alpha).$$

Since $e^u < 1.1u + 1$ for $0 < u < 0.1$, by Theorem 3, we have

$$\Xi \leq (1-p) \cdot \left(1.1 \cdot (\alpha - 1) \cdot \frac{1.09\alpha + 0.91}{2} \cdot \frac{(\lfloor x_{n+1} \rfloor)^2}{2n\lambda} + 1\right)$$
$$+ p \cdot \left(1.1 \cdot (\alpha - 1) \cdot \frac{1.09\alpha + 0.91}{2} \cdot \frac{(\lfloor x_{n+1} \rfloor + 1)^2}{2n\lambda} + 1\right).$$

Since $1 + u < e^u$, and $\alpha > 1$

$$\Xi \leq (\alpha - 1) \cdot \frac{1.199\alpha + 1.001}{4n\lambda}(x_{n+1}^2 + p - p^2) + 1$$
$$\leq \exp\left((\alpha - 1) \cdot \frac{1.2\alpha + 1}{4n\lambda}(x_{n+1}^2 + p - p^2)\right).$$

Hence we have $B_\alpha = \frac{1}{\alpha - 1} \ln \Xi \leq \frac{1.2\alpha + 1}{4n\lambda}(x_{n+1}^2 + p - p^2)$. □

### C.3 Proof of Lemma 4

PROOF. We sketch the proof for the first inequality. The proof for the second inequality is similar. We first examine the output distribution of 1SMM(X), which is a mixture of shifted symmetric Skellam distributions. In particular, there are $2^n$ shifted Skellam distributions, each of which corresponds to a sequence of $n$ Bernoulli trials, performed by all $n$ clients. For example, let us consider the event that all clients succeed in the Bernoulli trials. The probability of this event is $\prod_{i=1}^n (x_i - \lfloor x_i \rfloor)$. Conditioned on this, the output distribution is $\sum_{i=1}^n \lceil x_i \rceil + Sk(\lambda, \lambda)$. In general, we denote the weight of the shifted Skellam distribution as $w_j$ and the corresponding shift as $s_j$, $j = 1, \ldots, 2^n$. Then, 1SMM(X) is distributed as $s_j + Sk(\lambda, \lambda)$, with probability $w_j$.

Now, let us consider the neighboring dataset $X'$, obtained by adding one tuple $x_{n+1}$ to $X$, i.e., $X' = X \cup \{x_{n+1}\}$. Observe that the output distribution of 1SMM($X'$) is still a mixture of shifted Skellam distributions. The difference is that for $X'$, there are $2^{n+1}$ shifted Skellam distributions, each of which corresponds to a sequence of $(n+1)$ Bernoulli trials, performed by all $(n+1)$ clients. Now let us fix the outcome of the Bernoulli trials performed by the first $n$ clients (i.e., the clients that are both in dataset $X$ and $X'$). Then for each outcome, adding the $(n+1)$-th client introduces two more possibilities. Namely, with probability $x_{n+1} - \lfloor x_{n+1} \rfloor$, the shifted vector $s_j$ is further shifted by $\lceil x_{n+1} \rceil$, and otherwise by $\lfloor x_{n+1} \rfloor$. Then 1SMM($X'$) is distributed as $s_j + \lceil x_{n+1} \rceil + Sk(\lambda, \lambda)$, with probability $w_j \cdot (x_{n+1} - \lfloor x_{n+1} \rfloor)$; and $s_j + \lfloor x_{n+1} \rfloor + Sk(\lambda, \lambda)$, with probability $w_j \cdot (1 - x_{n+1} + \lfloor x_{n+1} \rfloor)$. Hence, we can also view the output distribution of 1SMM($X'$)) as a mixture of $2^n$ distributions, where each distribution itself is a mixture distribution. Then by Theorem 2 (Theorem 13 in Ref. [53]), we only need to consider the worst case Rényi divergence between one shifted Skellam distribution and a mixture of two shifted Skellam distributions. Let $p := x_{n+1} - \lfloor x_{n+1} \rfloor$. We have

$$D_\alpha(1\text{SMM}(X) \| 1\text{SMM}(X'))$$
$$\leq \max_{s_j} D_\alpha(s_j + Sk(\lambda, \lambda) \|$$
$$p \cdot (s_j + \lceil x_{n+1} \rceil + Sk(\lambda, \lambda)) + (1-p) \cdot (s_j + \lfloor x_{n+1} \rfloor + Sk(\lambda, \lambda)))$$
$$= D_\alpha(0 + Sk(\lambda, \lambda) \|$$
$$p \cdot (0 + \lceil x_{n+1} \rceil + Sk(\lambda, \lambda)) + (1-p) \cdot (0 + \lfloor x_{n+1} \rfloor + Sk(\lambda, \lambda)))$$
$$= D_\alpha(1\text{SMM}(\{0, \ldots, 0\}) \| 1\text{SMM}(\{0, \ldots, 0, x_{n+1}\})). \quad \square$$

### C.4 Proof of Lemma 5

PROOF. Let $p_k = \exp(-2\lambda)I_{|k|}(2\lambda), k = 0, \pm 1, \pm 2, \ldots, x := x_{n+1}$, $p := x - \lfloor x \rfloor$, and $s = \lfloor x \rfloor$. Then we have

$$\exp((\alpha - 1)B_\alpha)$$
$$= \sum_{k=-\infty}^{\infty} \left(\frac{(1-p) \cdot p_{k-s} + p \cdot p_{k-s-1}}{p_k}\right)^{\alpha - 1}$$
$$\cdot ((1-p) \cdot p_{k-s} + p \cdot p_{k-s-1})$$
$$= \sum_{k=-\infty}^{\infty} \left(\frac{(1-p) \cdot p_{k-s} + p \cdot p_{k-s-1}}{p_k}\right)^{\alpha} \cdot p_k$$
$$\leq \sum_{k=-\infty}^{\infty} \left((1-p) \cdot \left(\frac{p_{k-s}}{p_k}\right)^{\alpha} + p \cdot \left(\frac{p_{k-s-1}}{p_k}\right)^{\alpha}\right) \cdot p_k$$
$$= (1-p) \cdot \sum_{k=-\infty}^{\infty} \left(\frac{p_{k-s}}{p_k}\right)^{\alpha} \cdot p_k + p \cdot \sum_{k=-\infty}^{\infty} \left(\frac{p_{k-s-1}}{p_k}\right)^{\alpha} \cdot p_k.$$

The inequality follows from the fact that for $a, b > 0, \alpha > 1$ the function

$$f(p) := ((1-p) \cdot a + p \cdot b)^\alpha - ((1-p) \cdot a^\alpha + p \cdot b^\alpha)$$

is convex for $p \in [0, 1]$ and the fact that $f(0) = f(1) = 0$. The proof for $A_\alpha$ follows from Theorem 1. □

### C.5 Proof of Corollary 2

PROOF. By linearity of expectation, the error at each dimension is composed of the error due to the $n$ independent Bernoulli trials, and the $n$ independent Skellam noises sampled from $Sk(\lambda, \lambda)$. The proof then follows from Theorem 5 and the fact that adding $n$ independent Skellam noise sampled from $Sk(\lambda, \lambda)$ results in a Skellam noise sampled from $Sk(n\lambda, n\lambda)$. □

### C.6 Proof of Theorem 6

PROOF. We first apply the subsampling lemma of RDP (Lemma 2) on $d$-dimensional SMM (Corollary 1), and obtain the privacy cost of each iteration as follows: $\tau_1 = \frac{1}{\alpha - 1} \cdot \log\left((1-q)^{\alpha-1}(\alpha q - q - 1) + \sum_{l=2}^{\alpha} \binom{\alpha}{l}(1-q)^{\alpha-l}q^l e^{(l-1)\tau(l)}\right)$. Then, the case for $T$ iterations follows from the composition of RDP (Lemma 1). □

### C.7 Proof of Theorem 8

PROOF OF THEOREM 8. We sketch the proof for Theorem 8. Similar to 1SMM, we let $p := x_{n+1} - \lfloor x_{n+1} \rfloor$. It suffices to bound the

following two terms:
$$X_\alpha := (1-p) \cdot Q_\alpha^{n,\sigma}(\lfloor x_{n+1} \rfloor)$$
$$+ p \cdot Q_\alpha^{n,\sigma}(\lceil x_{n+1} \rceil), \text{ and}$$
$$Y_\alpha := \frac{1}{\alpha-1} \cdot \ln\Big((1-p) \cdot \exp\big((\alpha-1) \cdot Q_\alpha^{n,\sigma}(\lfloor x_{n+1} \rfloor)\big)$$
$$+ p \cdot \exp\big((\alpha-1) \cdot Q_\alpha^{n,\sigma}(\lceil x_{n+1} \rceil)\big)\Big),$$

where we define $Q_\alpha^{n,\sigma}(s) := D_\alpha(s + Z_{n,\sigma^2} \| Z_{n,\sigma^2})$. For $Y_\alpha$, we have
$$Y_\alpha \le \frac{1}{\alpha-1} \cdot \ln\Big((1-p) \cdot \big(1.1(\alpha-1) \cdot Q_\alpha^{n,\sigma}(\lfloor x_{n+1} \rfloor) + 1\big)$$
$$+ p \cdot \big(1.1(\alpha-1) \cdot Q_\alpha^{n,\sigma}(\lceil x_{n+1} \rceil) + 1\big)\Big)$$
$$\le 1.1(1-p)Q_\alpha^{n,\sigma}(\lfloor x_{n+1} \rfloor) + 1.1 p Q_\alpha^{n,\sigma}(\lceil x_{n+1} \rceil)$$
$$= 1.1 X_\alpha,$$

where the first inequality is because $e^u < 1.1u + 1$ for $0 < u < 0.1$; and the second inequality is because $\ln(1+u) < u$ for $u > 0$. Then for $X_\alpha$, we have
$$X_\alpha \le (1-p)\frac{\alpha(\lfloor x_{n+1} \rfloor)^2}{2n\sigma^2} + p\frac{\alpha(\lfloor x_{n+1} \rfloor+1)^2}{2n\sigma^2} + \tau_n, \text{ and}$$
$$X_\alpha \le (1-p)\frac{\alpha(\lfloor x_{n+1} \rfloor)^2}{2n\sigma^2} + p\frac{\alpha(\lfloor x_{n+1} \rfloor+1)^2}{2n\sigma^2} + \frac{\alpha x_{n+1}}{\sqrt{n}\sigma}\tau_n + \tau_n^2.$$

Hence,
$$X_\alpha \le \min\left\{\frac{\alpha c}{2n\sigma^2} + \tau_n, \frac{\alpha c}{2n\sigma^2} + \frac{\alpha \Delta_\infty}{\sqrt{n}\sigma}\tau_n + \tau_n^2\right\}$$
□

## Appendix D USEFUL LEMMATA

LEMMA 6. Let $Q_{s,t}(v) = \prod_{i=1}^s \frac{v+i+\sqrt{(v+i)^2+t^2}}{t}$. Then,
$$\frac{I_{|v|}(t)}{I_{|v+s|}(t)} \le Q_{s,t}(v).$$

PROOF. By properties of the modified Bessel function of the first kind [6], we have inequality $\frac{I_{v-1}(t)}{I_v(t)} < \frac{v+\sqrt{v^2+t^2}}{t}$ for $v \ge 0$, and recurrence relation $I_{v+1}(t) = I_{v-1}(t) - \frac{2v}{t}I_v(t)$, which leads to
$$\frac{I_{v+1}(t)}{I_v(t)} < \frac{-v+\sqrt{v^2+t^2}}{t}, \quad v \ge 0.$$
When $v < 0$, substituting $v$ for $-v-1$ in the previous formula,
$$\frac{I_{|v|}(t)}{I_{|v+1|}(t)} = \frac{I_{-v}(t)}{I_{-v-1}(t)} < \frac{v+1+\sqrt{(v+1)^2+t^2}}{t}$$
When $v \ge 0$, we also have
$$\frac{I_{|v|}(t)}{I_{|v+1|}(t)} = \frac{I_v(t)}{I_{v+1}(t)} < \frac{v+1+\sqrt{(v+1)^2+t^2}}{t}$$
Therefore, for every $v$,
$$\frac{I_{|v|}(t)}{I_{|v+s|}(t)} < \prod_{i=1}^s \frac{v+i+\sqrt{(v+i)^2+t^2}}{t} := Q_{s,t}(v). \quad \square$$

LEMMA 7. For any $w \ge 0$,
$$w + \sqrt{w^2+1} \le e^w.$$

PROOF. The inequality holds when $w = 0$. By taking the derivative, it can be verified that the function $e^w - (w + \sqrt{w^2+1})$ is increasing with respect to $w$. □

LEMMA 8. For any $0 < a \le w$,
$$\frac{w+\sqrt{w^2+1}}{(a-w)+\sqrt{(a-w)^2+1}} \le e^{2w-a}.$$

PROOF. By Lemma 7, $w + \sqrt{w^2+1} \le e^w$. Then, by Lemma 7 again,
$$\frac{1}{(a-w)+\sqrt{(a-w)^2+1}} = (w-a) + \sqrt{(w-a)^2+1} \le e^{w-a}. \quad \square$$

LEMMA 9. For any $0 < a \le w$,
$$(w+\sqrt{w^2+1}) \cdot ((a-w)+\sqrt{(a-w)^2+1}) \le e^a.$$

PROOF. By Lemma 7, the inequality holds when $w = a$. By taking derivative, it can be verified that the left hand side is decreasing with respect to $w$. □

LEMMA 10. When $-s < z < s$, we have
$$Q_{s,2\lambda}(z) \le \exp\left(\frac{s(s+z)}{2\lambda}\right).$$

PROOF. Use Lemma 7. When $0 \le z < s$,
$$Q_{s,2\lambda}(z) \le \prod_{i=1}^s \exp\left(\frac{z+i}{2\lambda}\right) = \exp\left(\frac{s(s+1+2z)}{4\lambda}\right) \le \exp\left(\frac{s(s+z)}{2\lambda}\right).$$
When $-\frac{s}{2} \le z < 0$, using the equality
$$\frac{b+\sqrt{b^2+(2\lambda)^2}}{2\lambda} \cdot \frac{-b+\sqrt{(-b)^2+(2\lambda)^2}}{2\lambda} = 1,$$
we have
$$Q_{s,2\lambda}(z) = \prod_{i=1}^s \frac{z+i+\sqrt{(z+i)^2+(2\lambda)^2}}{2\lambda}$$
$$= \prod_{i=-2z}^s \frac{z+i+\sqrt{(z+i)^2+(2\lambda)^2}}{2\lambda}$$
$$\le \prod_{i=-2z}^s \exp\left(\frac{z+i}{2\lambda}\right)$$
$$= \exp\left(\frac{s(s+2z-1)}{4\lambda}\right)$$
$$\le \exp\left(\frac{s(s+z)}{2\lambda}\right).$$
When $-s < z < -\frac{s}{2}$, we have
$$Q_{s,2\lambda}(z) = \prod_{i=1}^{-2z-s-1} \frac{z+i+\sqrt{(z+i)^2+(2\lambda)^2}}{2\lambda}$$
$$\le 1 \le \exp\left(\frac{s(s+z)}{2\lambda}\right).$$
□